\documentclass{article}

\usepackage{arxiv}

\usepackage[utf8]{inputenc} 
\usepackage[T1]{fontenc}    
\usepackage{hyperref}       
\usepackage{url}            
\usepackage{booktabs}       
\usepackage{amsfonts}       
\usepackage{nicefrac}       
\usepackage{lipsum}

\usepackage{enumerate}

\usepackage{amsmath}
\usepackage{setspace}
\usepackage{mathtools}
\usepackage{accents}
\usepackage{amssymb}
\usepackage{enumitem}
\usepackage{hyperref}
\usepackage{enumitem}
\usepackage{longtable}
\usepackage{fixmath}
\usepackage{wrapfig}
\usepackage{color}
\usepackage{algorithm}
\usepackage{algpseudocode}
\usepackage{comment}
\usepackage{subfig}
\usepackage{soul}
\usepackage{lineno}
\usepackage{graphicx,multicol}
\usepackage{url}
\usepackage{hyperref}
\usepackage{multicol,multirow}
\usepackage[dvipsnames]{xcolor}
\usepackage{tikz} 
\usepackage{subcaption}

\title{Multi-Layer Attention-Based Explainability via Transformers for Tabular Data}

\author{
  Andrea Trevi\~no Gavito\\
  Department of Industrial Engineering and Management Sciences\\
  Northwestern University\\
  Evanston, IL 60208 \\
  \texttt{andrea.tg@u.northwestern.edu} \\
   \And
   Diego Klabjan \\
  Department of Industrial Engineering and Management Sciences\\
  Northwestern University\\
  Evanston, IL 60208 \\
  \texttt{d-klabjan@northwestern.edu} \\
   \And
   Jean Utke \\ 
  Data, Discovery and Decision Science \\
  Allstate Insurance Company\\
  Northbrook, IL 60062 \\
  \texttt{jutke@allstate.com} \\
}
\date{today}

\begin{document}

\maketitle

\begin{abstract}
We propose a graph-oriented attention-based explainability method for tabular data. Tasks involving tabular data have been solved mostly using traditional tree-based machine learning models which have the challenges of feature selection and engineering. With that in mind, we consider a transformer architecture for tabular data, which is amenable to explainability, and present a novel way to leverage self-attention mechanism to provide explanations by taking into account the attention matrices of all heads and layers as a whole. The matrices are mapped to a graph structure where groups of features correspond to nodes and attention values to arcs. By finding the maximum probability paths in the graph, we identify groups of features providing larger contributions to explain the model's predictions. To assess the quality of multi-layer attention-based explanations, we compare them with popular attention-, gradient-, and perturbation-based explanability methods.
    
\end{abstract}

\section{Introduction}

Tabular data (TD) is one of the most prevalent data modalities in real world applications. It is widely used in critical daily-life fields such as medicine, transportation, insurance and finance, many times in combination with additional unstructured data modalities like images, videos, and text. Due to its ubiquity and relevance, it is becoming increasingly common to allow automated systems to use this data as input to models trained to propose or even directly make decisions. Nonetheless, the models are often used without a clear understanding of how and why model decisions come about \cite{confalonieri2021, tjoa2019}. Given the nature of the application areas, these decisions may have significant impact and consequences in human lives, highlighting an accentuated need for interpretable and explainable machine learning models. Understanding the rationale behind the decision making of intelligent systems remains the most pertinent factor towards building trust in Artificial Intelligence and further enabling its usage \cite{das2020, gunning2019}. 

The huge advances of modern deep learning (DL) models have focused mostly on unstructured data sources \cite{Goodfellow-et-al-2016, Roy2019}, as have most of the efforts made towards generating better prediction explanations and interpretable models. For TD, traditional machine learning models such as boosting and random forests continue to be de-facto choices \cite{tjoa2019}. Yet, despite their good performance, models based on tree ensembles are usually not as interpretable as some simpler -but less successful- models like logistic regression or single decision trees \cite{caruana2015}. In this sense, improving DL models for TD would allow both single- and multi-modal problems with TD to benefit from more widely explored explainable models. Additionally, DL models leverage gradient descent learning, which is not supported by tree-based models. This type of optimization decreases the need for feature selection and engineering and facilitates joint training of models taking inputs from multiple data sources with mixed modalities \cite{ tabnet2019, fttransf2021}. 

In particular, a set of models that have shown promising results at pushing the boundaries of DL for TD are transformers \cite{tabnet2019, fttransf2021,tabtransf2020,saint2021, autoint2019}. Unlike other DL models, transformers have not only obtained great success in diverse tasks, but also possess a built-in capability to provide explanations for its results via attention \cite{attention2017}. Following this lead, we propose multi-layer attention-based explainability\footnote{The extent to which the attention mechanism should be used for providing explanations is a matter of interest and an ongoing debate \cite{jain,thorne,serrano,wiegreffe,dong}. However, this discussion is contingent upon the definition of the term \textit{explainability}. In the context of this paper, we adhere to Wiegreffe and Pinter\cite{wiegreffe}'s terminology and define it as ``providing a plausible, but not necessarily faithful rationale for model prediction".} via transformers for TD, a novel method that leverages attention mechanism and combines it with knowldedge transfer and graph concepts to enable a better understanding of how groups of tabular features influence the transformer's decisions. To this end, a transformer model is trained on a classification task using TD as input. In tabular inputs, it is common to have several features that represent similar underlying concepts. As the number of features increases, their collective importance might end up being diluted across the relative importance of a large number of single features, making it hard to pinpoint relevant explanations at a conceptual level. To account for this, instead of assigning importance or relevance to each individual feature, meaningful groups of features are created a priori. A transformer model is trained with these groups as inputs and the most relevant concepts for the classification task are identified by the model at the conceptual or group level. We train transformers on three datasets and generate multi-layer attention-based explanations for their prediction, i.e., we identify groups of features that have the largest impact on the model's decision. For a transformer with a single head, prior work considers attention only at the last layer, which disregards information of all preceding layers. Our methodology considers all layers. To cope with the assumption of a single head, we use knowledge distillation \cite{hinton2015distilling} to train a single-head but multi-layer transformer based on a trained multi-head transformer and apply graph-based explainability on the student. We further compare our explanations with those provided by other widely known explainability methods.

In summary, the contributions of this work are as follows.
\begin{enumerate}
    \item We investigate explainable models based on transformers for tabular data.
    \item We propose a graph-oriented attention-based explainability method via transformers for tabular data.
    \item We introduce an attention-based explainability method that accounts for all layers of multi-head transformer models.
    \item We compare this approach to attention-, gradient-, and perturbation-based explainability methods.

\end{enumerate}

The rest of this paper is organized as follows. In Section \ref{sec:lr}, the related work is discussed. Section \ref{sec:model} describes the proposed model: the conceptual transformer model for TD in Section \ref{sec:transf} and the explainability method used to identify relevant concepts in Section \ref{sec:alg}. Section \ref{sec:comp} provides the computational study, experimental details, results, and visualizations. Conclusions are given in Section \ref{sec:conclusion}.

\section{Related work} \label{sec:lr}
\subsection{Explainability for Deep Learning}

The field of explainable Artificial Intelligence (XAI) has received increasing interest over the past decade. Surveys, reviews, and articles such as \cite{confalonieri2021, das2020, islam, samek2019explainable, tjoa2019, vilone, notions,   visualsurvey} have synthesized its main motivations, approaches, and challenges. 

In a broad sense, XAI algorithms for DL can be organized into three major groups: perturbation-based, gradient-based, and, more recently, attention-based. Within the most famous perturbation-based methods, we find LIME \cite{LIME}, which generates a local approximation for a given model around a specific prediction, and SHAP \cite{SHAP}, which measures a feature's importance as the change in the expected prediction when conditioning on it. These methods are model agnostic, but have often been applied in DL settings. On the other hand, gradient-based algorithms have focused on DL algorithms, as they leverage gradient information to assess the relevance of the model's inputs to make its decision. The classic gradient-based methods are saliency maps \cite{Simonyan}, used for explaining the predictions of convolutional neural networks. In saliency maps, the gradients of the predictor function with respect to the input are computed and used to identify parts of the image that contribute the most either to the final decision or to a specific layer in the network. Another example is Grad-CAM \cite{gradcam}, in which gradients are used to compute an importance score that allows class-specific neuron activity visualization in images (referred to as activation maps). A more general gradient-based method is layer-wise relevance propagation \cite{LWRP}, while other well known methods falling under this category are Deeplift \cite{deeplift} and SmoothGrad \cite{smoothgrad}.

Attention-based explanations gained relevance along with the success of transformer models \cite{attention2017} in a variety of application areas such as natural language processing, computer vision, and speech processing \cite{transformersurvey}. Transformers possess a built-in XAI method: the attention mechanism, which generates probability distributions over features and further interprets them as feature importances or contributions. Various forms of attention-based explainability have been investigated in the past, and determining the most effective method for obtaining meaningful explanations from attention heads remains a topic of ongoing research. Attention matrix aggregation has been conducted through the utilization of a variety of operations, including mean, multiplication, and extraction of the $i$'th and maximum element at the head, layer, and matrix levels \cite{mylonas,lopardo}. Nevertheless, the extent to which meaningful information is compromised by the aggregation remains unclear.

Attention has been further combined with other attributes to generate explanations. For instance, in \cite{voita}, layer-wise relevance propagation is applied to transformers, and  \cite{chefer} builds on that idea by including gradient information into the explanations. A reinforcement learning-based method for perturbation sampling was proposed as a means of identifying key features. This method involves masking inputs and observing the resultant impact on attention matrices \cite{niu}. Directed acyclic graphs have been utilized for explainability purposes as well \cite{abnar}, but aggregation of attention matrices across the heads at each layer by averaging is performed. In contrast, our proposed method accounts for all head's information and transfers their knowledge into a distilled model.
Transformers were initially introduced for machine translation, but have been extended and customized for diverse tasks such as object detection \cite{detr} and, more recently to multimodal settings \cite{vilbert, lxmert}. Yet, the emphasis of the work reviewed in this section is on tasks concerning vision and language rather than on TD.
 
\subsection{Transformers for Tabular Data}
Following the success on unstructured data, transformers have also proven to have good performance on TD. 
One of the first models to leverage transformers for TD is TabNet  \cite{tabnet2019}, which adopts transformer blocks to mimic the structure of decision trees and incorporates sequential attention to select which features to focus on at each step. Similarily, SAINT \cite{saint2021} combines self-attention with inter-sample attention to attend both rows and columns in the TD. TabTransformer \cite{tabtransf2020} uses self-attention transformers to map categorical features to conceptual embeddings. Continuous features are not passed through the transformer architecture, but concatenated with its output for further processing. FT-Transformer \cite{fttransf2021} introduces a feature tokenizer to adapt the transformer architecture to TD. While all of these methods use attentive transformers on TD, none of them consider multiple layers for explainability of attention or incorporate a priori conceptual information to the architecture. In the context of tabular data, graphs come into consideration in graph attention networks \cite{velickovic2018graph}, neural network architectures that operate on graph-structured data. However, to the best of our knowledge, no further work has been done on leveraging graphs for attention matrices for TD.

\section{Proposed Model} \label{sec:model}

In this section, we introduce multi-layer attention-based explainability leveraging transformers for TD. Single-head transformers are more amenable to explanations. However, multi-head attention allows transformer models to focus on different parts of the inputs simultaneously. Each attention head can attend to different aspects of the input, leading to richer representations. In order to obtain the benefits of multi-head multi-layer transformers while simultaneously enabling 
 high-quality explanations with minimal information loss, we use knowledge distillation (student-teacher paradigm) to train a single-head but multi-layer transformer based on a trained multi-head transformer. We propose the following process. First, a multi-head transformer (referred to as the teacher transformer) is trained. Then, a single-head transformer (the student transformer) is trained based on the output predictions of the teacher. Finally, explanations are extracted from the student by using attention values from all layers, containing distilled information from every head and layer of its corresponding teacher. In the following subsections, we describe how the transformer architecture is adapted to account for the specific structure of TD and incorporate a priori conceptual information. Next, we describe how we map the underlying self-attention mechanism into attention graphs.

\subsection{Contextual Transformer Encoder for TD}  \label{sec:transf}
The original transformer model was designed for sequence transduction tasks on text data. TD and text have inherently different structures and as such, their feature engineering strategies differ as well. For instance, preprocessing raw text data is usually done by tokenization, whereas for TD, it is common to normalize numerical features and one-hot-encode categorical. 

To account for these differences, two main changes are made to the transformer architecture. First, groups of features representing conceptual information are manually defined before training. Hence, for the TD case, instead of having attention matrices where each word's projection attends every other word's projection, we have conceptual groups of features that attend other groups of features. Features within TD can often be grouped intuitively and naturally based on factors such as their source (e.g. sensors, monitoring systems, surveys) and type (e.g demographic, ordinal, or geospatial data). Unlike other data modalities, in the context of TD, this type of a-priori knowledge is generally accessible. In this sense, conceptual grouping exploits an advantage specific to TD that we aim at accommodating for in our design\footnote{In the absence of such prior information, concept groups can be easily formed through clustering or, with minor adjustments, the individual features can be utilized as direct inputs to our model.}. Second, given that TD does not provide sequential information, positional encoding is disabled. The adapted transformer architecture is trained for the classification task at hand.

Let $x_{1}  \in \mathcal{R}^{k_{1}}$, $x_{2}  \in \mathcal{R}^{k_{2}}$, ..., $x_{m}  \in \mathcal{R}^{k_{m}}$ be the concept groups of features. We project $x_{i}$ into latent space $\mathcal{R}^{d}$ by defining: $\tilde{x_{i}}= D_{i}x_{i} \in \mathcal{R}^{d}$, with $D_{i} \in \mathcal{R}^{d\times k_{i}}$ trainable. Then, $X = [\tilde{x_{1}}, ..., \tilde{x_{m}}]^{T} \in \mathcal{R}^{m\times d}$. Following \cite{attention2017}, we obtain attention coefficients $a_{i,j}$ by defining $V = XW^{V}$, $K=XW^{K}$, $Q=XW^{Q}$, with $W^{V}, W^{K}, W^{Q} \in \mathcal{R}^{d\times d}$ trainable matrices, and $V, K, Q \in \mathcal{R}^{m\times d}$. We have that $\frac{QK^{T}}{\sqrt{d}} \in \mathcal{R}^{m\times m}$, $A = [a_{i,j}] = softmax(\frac{QK^{T}}{\sqrt{d}}) \in \mathcal{R}^{m\times m} $ and $AV \in \mathcal{R}^{m\times d}$.

The above-mentioned transformer encoder will then have $N  \times  h$ attention matrices, where $N$ is the number of encoder layers and $h$ is the number of attention heads. While $N$ and $h$ are tunable hyperparameters, for optimal results they are almost always larger than 1. For XAI purposes, as $N$ and $h$ increase, multi-head attention becomes harder to interpret. To leverage the strengths of multi-head attention while simultaneously prioritizing explainability, we use knowledge distillation \cite{hinton2015distilling} as a means to learn a simplified, single-head student transformer model that can generalize in a similar fashion as the original teacher model. The student architecture has $h = 1$ and $M$ encoder layers, where $M$ typically meets $M>N$. Furthermore, to improve the entropy of the attention matrices, a penalization term is added to the student's cross entropy loss function, yielding: 
\begin{center}
$L = - \sum\limits_{i = 1}^{n} y_{i}log(\hat y_{i}) + \lambda \sum\limits_{l=1}^{M} \sum\limits_{j,k = 1}^{m} a^{l}_{j,k} log(a^l_{j,k}) $  
\end{center}
where $n$ is the number of training samples, $\lambda$ is the penalization term hyperparameter,  $a^l_{j,k}$ is the value in the $j^{th}$ row and $k^{th}$ column of attention matrix $A^{l}$ corresponding to the $l^{th}$ encoder layer, and $y_{i}$ and $\hat y_{i}$ are the predictions of the multi-head teacher and the predicted value of the student for sample $i$, respectively.

\subsection{Multi-Layer Attention-Based Explainability} \label{sec:alg}

Multi-layer attention-based explainability for TD (MLA) leverages the conceptual transformer encoder's attention mechanism described in Section \ref{sec:transf} and maps the attention matrices across encoder layers into a directed acyclic graph (DAG). In the DAG, the vertices correspond to concept groups of features and the arcs to attention values. We further identify the concept group with the largest contribution to the prediction, that is, the \textit{best concept group} to explain the output, as the input group corresponding to the path of the maximum probability in the DAG. 

\begin{figure}[H]
    \centering
      \includegraphics[width=0.7\linewidth]{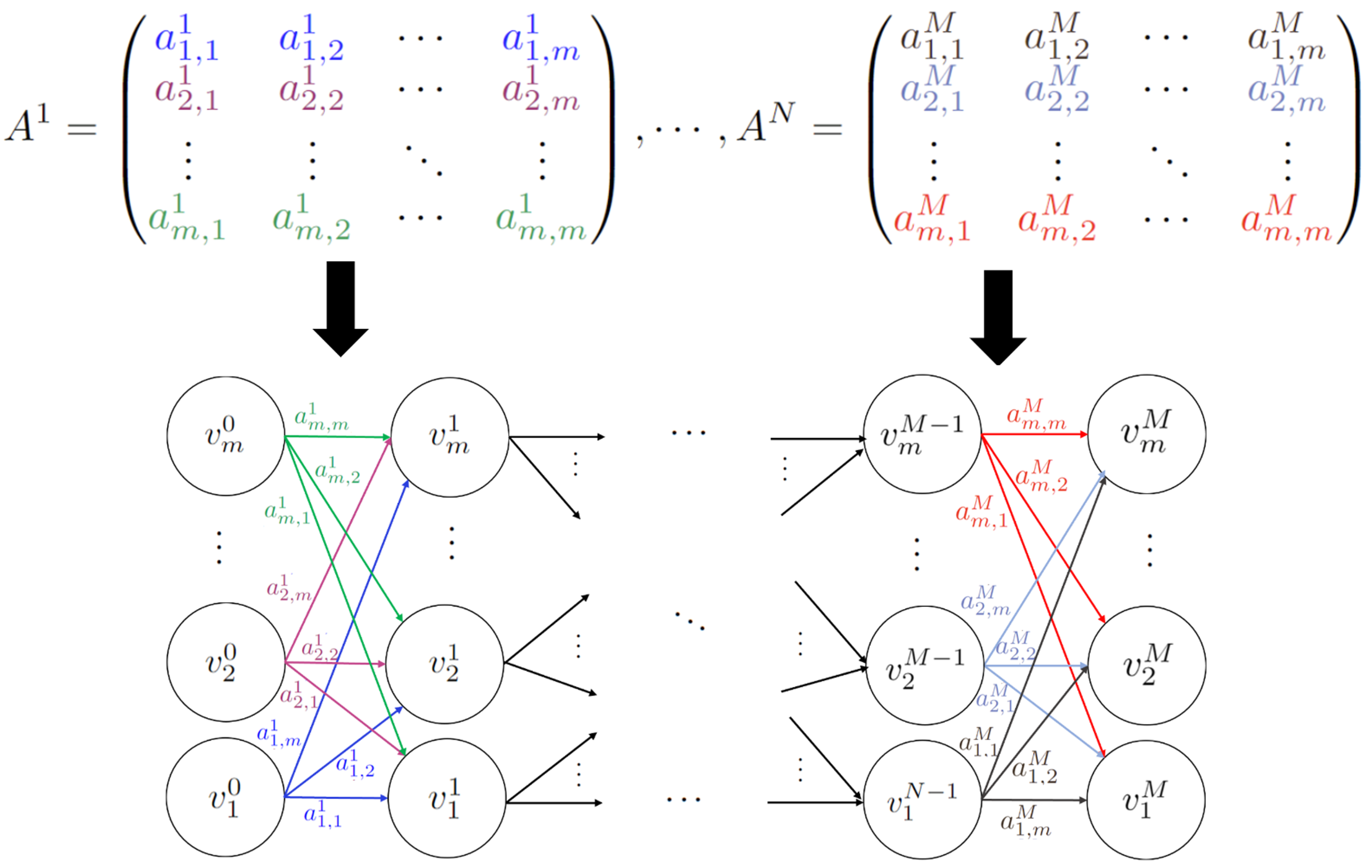}
    \caption{Graph $D= (V,A)$} \label{fig1} 
\end{figure}

For a given conceptual transformer, we have a collection of attention matrices $A^{l} = (a^{l}_{j,k})$ with $l \in \{1, ..., M\}$, and $j,k \in \{1, ..., m\}$ as described above. We define $D = (V, A)$ a weighted DAG as follows. Let  $V = \bigcup_{l=0}^{M}\{v^{l}_{c}\}$  and  $(v^{l-1}_{\hat c}, v^{l}_{\tilde{c}}) \in A$ ,
where arc $(v^{l-1}_{\hat c}, v^{l}_{\tilde{c}})$ has weight  $ a^{l}_{\hat{c},\tilde{c}}$, subscripts $\hat{c}$, $\tilde{c} \in \{1, ..., m\}$ correspond to concept groups, superscript $l$ corresponds to encoder layers, and $l = 0$ is a special case corresponding to the student's input layer. In Fig. \ref{fig1}, we present a visualization of the construction of $D$.

The maximum probability path $p$ is found using Dijkstra's algorithm \cite{dijkstra1959note}, and is of the form $p = \{v^{0}_{i_{0}},v^{1}_{i_{1}},...,v^{M}_{i_{M}}\}$ with arc cost of $-log( a^{l}_{j,k})$ for $a^{l}_{j,k}>0$, yielding path cost $-log \Big(\prod_{l = 1}^{M} a ^{l}_{i_{l-1}, i_{l} }\Big )$. Since we are particularly interested in the concept group corresponding to the most relevant input for the prediction, we focus on group $c = i_{0}$ corresponding to $v^{0}_{i_{0}}$.

Thus, we provide explanations to the student's predictions by finding the most relevant concept for the classification task, the \textit{best concept group}, defined as the concept group $c = i_{0}$ corresponding to the first vertex $v^{0}_{i_{0}}$ of the maximum probability path $p$ in graph $D$. Note that not always does a single concept group provide all the relevant information to make a prediction. To account for this, we rank additional concept groups iteratively. In each iteration we eliminate from the graph the starting point $v^{0}_{i_{0}}$ of the previously found highest probability path and then search for the respective next highest probability path in $D$. In our experiments, we use at most two \textit{best concept groups} to explain predictions.

\section{Computational study} \label{sec:comp}
\subsection{Datasets}
The proposed explainability model is tested on three datasets: UCI Forest CoverType \cite{Blackard:1999}, KDD’99 Network Intrusion dataset \cite{NI-ds}, and a real-world proprietary dataset as described below. These where selected due to their relatively large number of features (with a fair mix of numerical and categorical) and samples, allowing adequate conceptual aggregation.

\subsubsection*{Forest CoverType Dataset (CT)} 
In CT, the goal is to predict the most common cover type for each 30m by 30m patch of forest. We use the three most represented classes, resulting in approximately 425,000 training and 53,000 validation samples. The dataset consists of 10 quantitative features and two qualitative features, which were organized into the following five concept groups: 

\begin{itemize}
        \item \textit{Generals}: Elevation, aspect, and slope of the patch
        \item \textit{Distances}: Horizontal and vertical distances to hydrology\footnote{The dataset information file states that this refers to the "nearest surface water features."}, horizontal distances to roadways and fire points
        \item \textit{Hillshades}: Shades at 9am, noon, and 3pm
        \item \textit{Wild areas}: 4 different wilderness areas
        \item \textit{Soil types}: 40 different types of soil
\end{itemize}

\subsubsection*{Network Intrusion Dataset (NI)}
In NI, the classification task is to distinguishing between "bad" connections (intrusions or attacks) and "good" connections. Approximately 1,000,000 samples were used for training and almost 75,000 for validation, with each sample consisting of 53 features. The concept groups are defined following \cite{NI-ds-groups}:

\begin{itemize}
        \item \textit{Basic}: 20 features regarding individual TCP connections
        \item \textit{Content}: 14 features regarding the connection suggested by domain knowledge
        \item \textit{Traffic}: 9 features computed using a two-second time window
        \item \textit{Host}: 10 features designed to assess attacks which last for more than two seconds
\end{itemize}

\subsubsection*{Real-World Dataset (RW)}
The proprietary real-world dataset constitutes a binary classification problem. Tens of thousands of samples were used for training and validation. Each sample has approximately 100 features, which were subsequently arranged into 8 concept groups. Our access to this dataset has been granted for a restricted duration, resulting in its exclusion from certain experiments.

\subsection{Implementation and hyperparameters}
The experiments were implemented in Python and ran using GeForce RTX 2080 Ti GPU and Intel(R) Xeon(R) Silver 4214 CPU @ 2.20GHz for all datasets except RW, for which Tesla V100 GPU and Intel Xeon CPU E5-2697 v4 @2.30Hz were used.

The same hyperparameters were used for all teacher networks: $N=2$, $h=4$, $d=64$ and $128$ neurons in the internal layer. These parameters are standard choices for transformer encoders for TD; on the lower end for $N$ and $h$, and on the higher end for $d$ and neurons. The student's architecture is identical, but with $M=4$ and $h=1$. For training, we chose a dropout rate of $0.1$ to prevent overfitting while avoiding a large reduction of network's capacity. Additionally, we used a temperature of $2$, which provided a balance between producing reliable soft targets and avoiding to overly flatten the underlying probability distribution. A constant batch size of $128$ and the Adam \cite{adam} optimizer were employed. Between six and ten lambdas were tested for each dataset's training loss. The lambda corresponding to the highest F1 (for CT and NI) and accuracy (for RW) was selected for the final results, yielding $\lambda_{CT} = 0.005$, $\lambda_{NI} = 0.01$, and $\lambda_{RW} = 0.9$. To account for minibatch randomization, each experiment was repeated five times for each CT and NI student and ten times for each RW student, after which variance is already low (see Table \ref{f1}). In such cases, the \textit{best concept group} per method was defined as the mode of these experiments. However, the distributions of all repetitions are also presented in the pairwise method comparisons detailed in Section \ref{sec:exp}.

In order to assess the quality of multi-layer attention-based explanations, we first evaluate the performance of the conceptual transformer encoder for TD presented in Section \ref{sec:transf}. The model is compared against two go-to TD methods: LightGBM \cite{lightgbm} and XGBoost \cite{xgboost}, with 1,000 base learners each. Other DL and transformer approaches for TD were not considered in this comparison due to their higher computational requirements and similar (if not slightly worse) performance when compared to boosting methods (as reported in \cite{survey, fttransf2021}). For CT and NI, the aggregated results for five repetitions of each model are shown in Table \ref{f1}. As for RW, the conceptual transformer yielded a mean value of $0.88689$ with a standard deviation of $0.00234$. The teacher network's metrics are included for reference. Note that even though it outperforms all other models, our main goal is not to obtain the most performant model, but rather a comparable model that is better suited for explainability purposes. On the other hand, the conceptual transformer's performance displays comparable but not necessarily better absolute metrics than boosting models. However, unlike those, it does allow to obtain explanations without a significant loss in predictability. That is, conceptual transformers might not be the top-ranked classifier for all TD cases, but are able to provide explanations for their predictions. 

\begin{table}[ht]
\caption{Validation F1} \label{f1}
\begin{center}
\begin{tabular}{l|cc|cc|} 
& \multicolumn{2}{c|}{\textbf{CT}} & \multicolumn{2}{|c|}  {\textbf{NI}}  \\
\textbf{Models}  & \textbf{Mean} &\textbf{Std Dev}
&\textbf{Mean} &\textbf{Std Dev}\\
\hline 
    Conceptual Transformer & 0.96856 & 0.00055 & 0.88715 & 0.01537  \\
    LightGBM & 0.96208 & 0.00063 & 0.88875 & 0.00064  \\
    XGBoost & 0.97268 & N/A & 0.89226 & N/A \\
    Teacher Network & 0.9763 & N/A & 0.9076 & N/A\\
\end{tabular}
\end{center}
\end{table}

Having validated that its performance is satisfactory, the multi-layer attention-based explanations are extracted as discussed in Section \ref{sec:alg} and compared to those generated using the most popular method from each XAI group: attention-based, gradient-based and perturbation-based, as well as an additional attention-based method that, like MLA, utilizes DAGs, but aggregates information across the teacher network's attention heads through averaging.

\paragraph{Attention-based: Last-layer explainability (LL)}
We consider attention mechanism as presented in \cite{attention2017}. More specifically, the last layer's self-attention head of the student's encoder. The \textit{best concept group} to explain a given prediction is defined as that which corresponds to the highest attention value. 

\paragraph{Gradient-based: Saliency explainability (SA)}
In the same fashion as \cite{Simonyan}, but in the context of TD, the gradients of the loss function with respect to the input (concept groups) are computed. The \textit{best concept group} to explain a given prediction is defined as that which yields the largest mean absolute value.

\paragraph{Perturbation-based: Shapley additive explanations (SH)}
The SHAP \cite{SHAP} value of each feature is computed. The \textit{best concept group} is defined as that with the largest mean absolute SHAP value.

\paragraph{Attention head aggregation by averaging (AVG)} 
We consider the attention flow algorithm as presented in \cite{abnar}. In contrast to the other methods, this approach employs the teacher network instead of the distilled student and aggregates the heads from each layer by averaging them. The \textit{best concept group} to explain a given prediction is defined as in our proposed MLA model.

\subsection{Results} \label{sec:exp}
\subsubsection*{Explanation Distributions}
\begin{figure} [b]
    \centering
    \subfloat[\centering CT]{{\includegraphics[height=0.225\textwidth,width=0.33\linewidth]{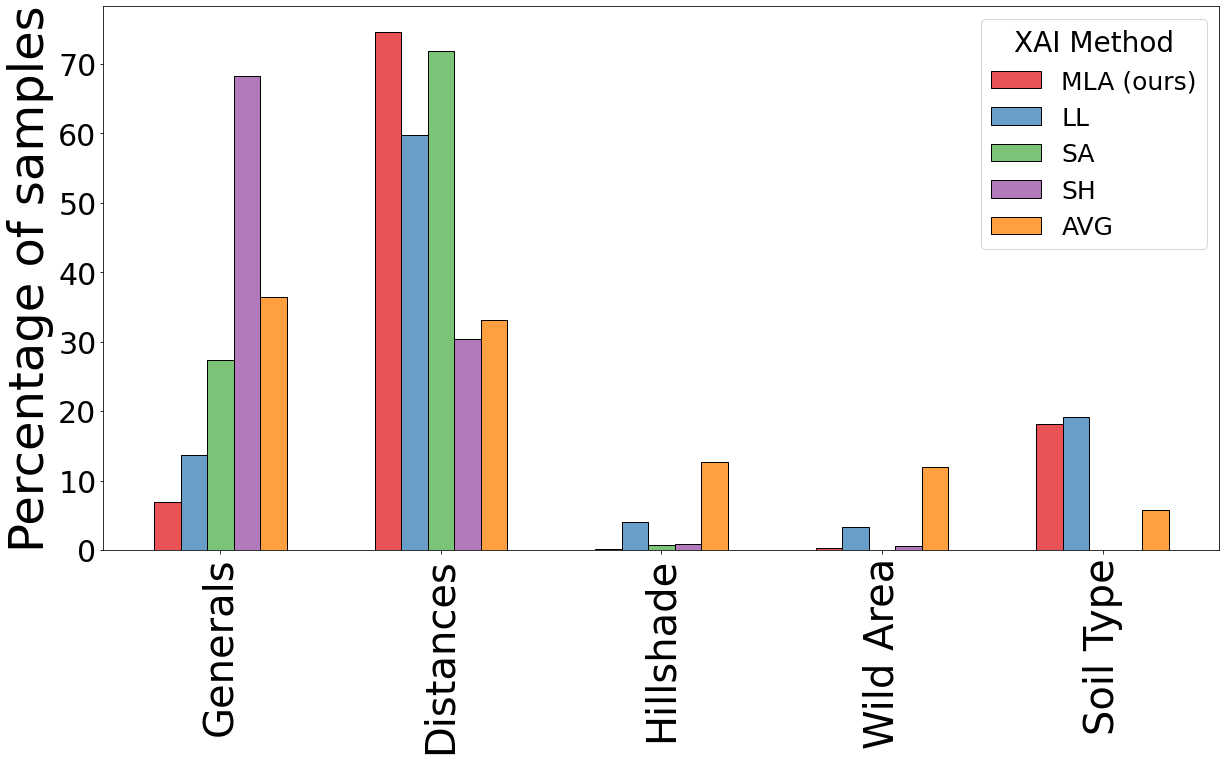} }}%
    \subfloat[\centering NI]{{\includegraphics[height=0.225\textwidth,width=0.33\linewidth]{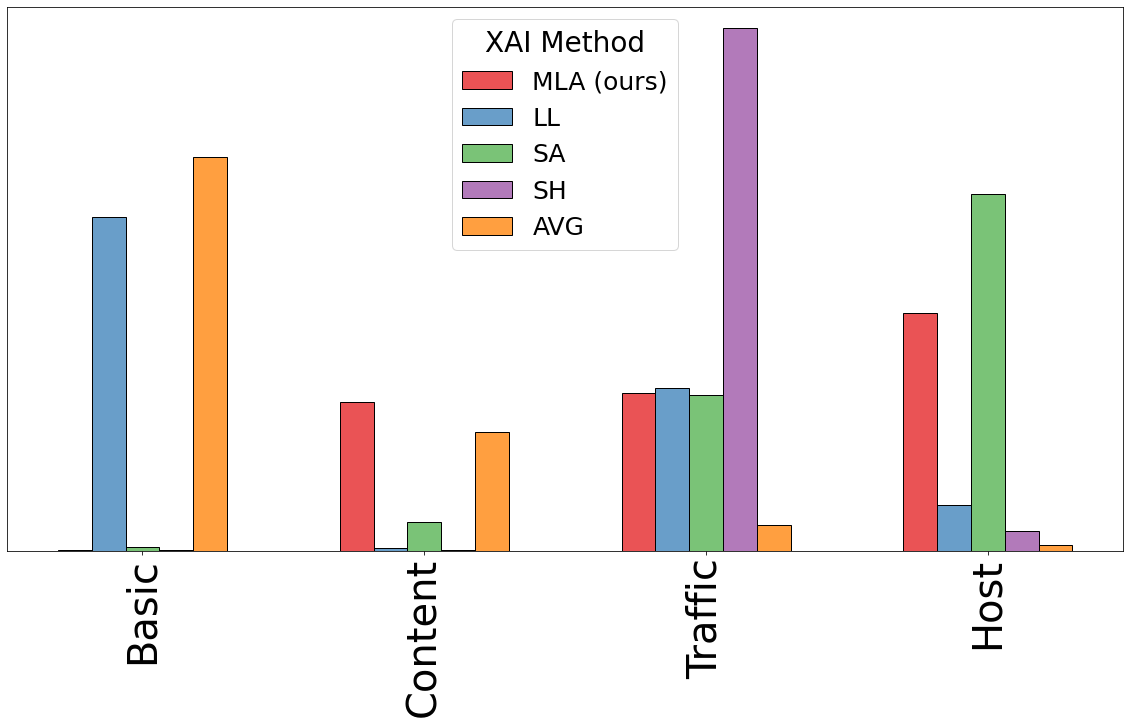} }}%
    \subfloat[\centering RW]{{\includegraphics[height=0.225\textwidth,width=0.33\linewidth]{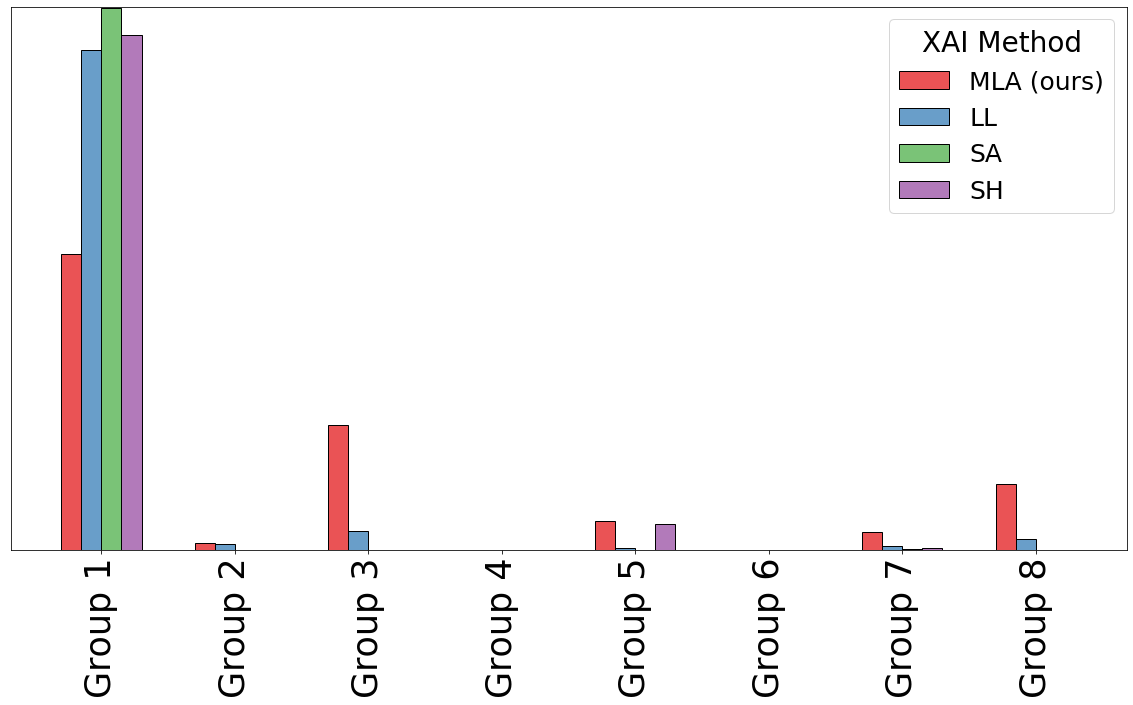} }}%
    \caption{Best concept group distribution per method}%
    \label{figBestCG}
\end{figure}

We analyze explanations at an aggregate level in Fig. \ref{figBestCG}, where the distributions of the \textit{best concept group} per method over the whole validation sets for CT and NI datasets are shown. For each type of explanation, we show the proportion of samples that deemed each concept group as best. In general, we do not distinguish between correctly and wrongly classified samples, unless explicitly stated. For each dataset, the number of incorrectly classified samples is less than $5\%$, which has no impact in the overall distributions (see Appendix A).

\begin{figure*}[!ht]
\centering
\includegraphics[width=\linewidth]{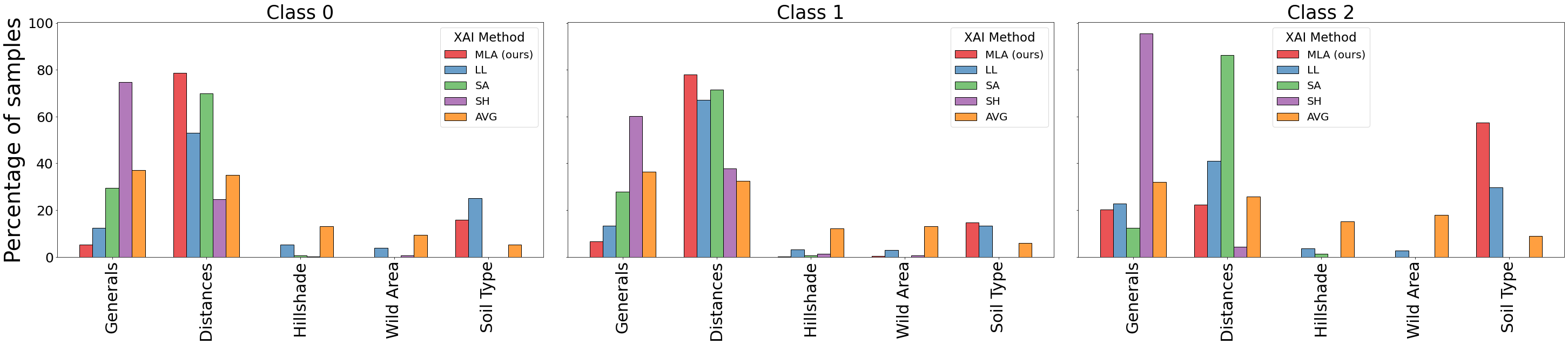}
\caption{CT's best group of features per method by class.}
\label{figBestCGperClass-CT}
\end{figure*}

\begin{figure*}[!ht]
\centering
\includegraphics[width=\linewidth]{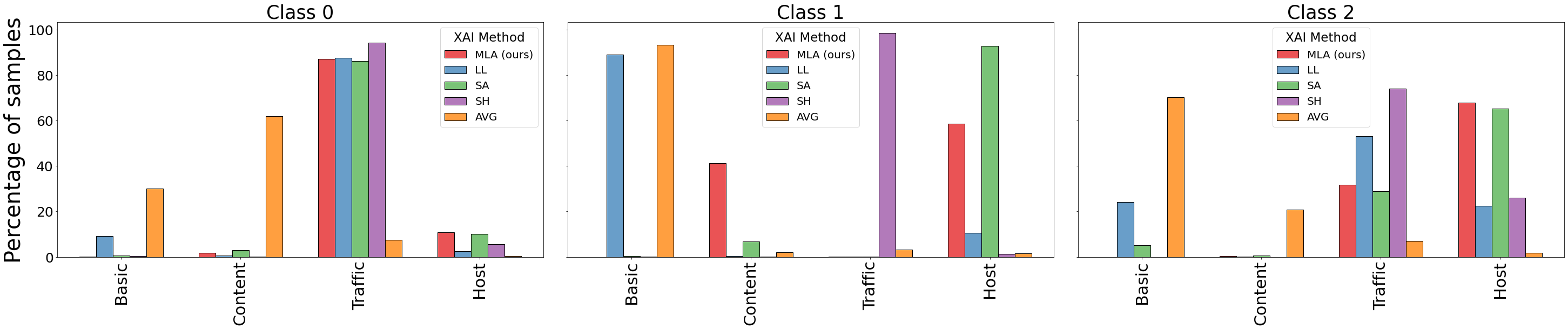}
\caption{NI's best group of features per method by class.}
\label{figBestCGperClass-NI}
\end{figure*}

In Fig. \ref{figBestCG}, we observe that SA and SH tend to focus on one or two concept groups to assign predictions, whereas LL, AVG, and MLA appear to take more groups into account when identifying differences among samples. This behavior is consistent across all datasets. In Fig. \ref{figBestCGperClass-CT} and Fig. \ref{figBestCGperClass-NI}, we zoom in and observe the above-mentioned distributions for CT and NI, respectively, but segmented by predicted class. 

For CT, a consistent focus on \textit{Generals} and \textit{Distances} is observed across all classes. However, LL and MLA seem to also assign large explainability values to \textit{Soil Type}, while AVG attends all groups to a certain extent. SA and SH exhibit a particular emphasis on a single concept group solely for Class 2. MLA's pattern is also altered for Class 2, with reduced focus on \textit{Distances} when compared to the other classes, and shifting it towards \textit{Soil Type}. LL displays a similar trend to MLA but less pronounced. Surprisingly, AVG demonstrates minimal to no alteration in its distribution when class control is applied.

In contrast, the methods show a stronger focus on specific groups for a given class for NI. All methods but AVG coincide in assigning the largest explainability value to \textit{Traffic} for Class $0$. Interestingly, MLA points at \textit{Content} and \textit{Host} as explanations to predict Class $1$, whereas LL and AVG point at \textit{Basic}, SA at \textit{Host}, and SH at \textit{Traffic}. Additionally, \textit{Host} and \textit{Traffic} are consistently referred to as the \textit{best concept groups} for Class $2$ by almost all groups (with some explanation value assigned to \textit{Basic} by LL as well). Again, the misaligned method is AVG, which points at \textit{Basic} and \textit{Content}.

In summary, AVG demonstrates the most distinct behavior among the methods, showing no clear alignment with any other method. Conversely, the remaining methods display significant consistency across each other on their \textit{best concept group} selection for CT, with a particularly strong alignment between LL and MLA. For NI, consistency is notable for Classes $0$ and $2$, with only Class $1$ revealing misalignments across methods. 

\begin{figure*}[!h]
\centering
\includegraphics[width=0.85\linewidth]{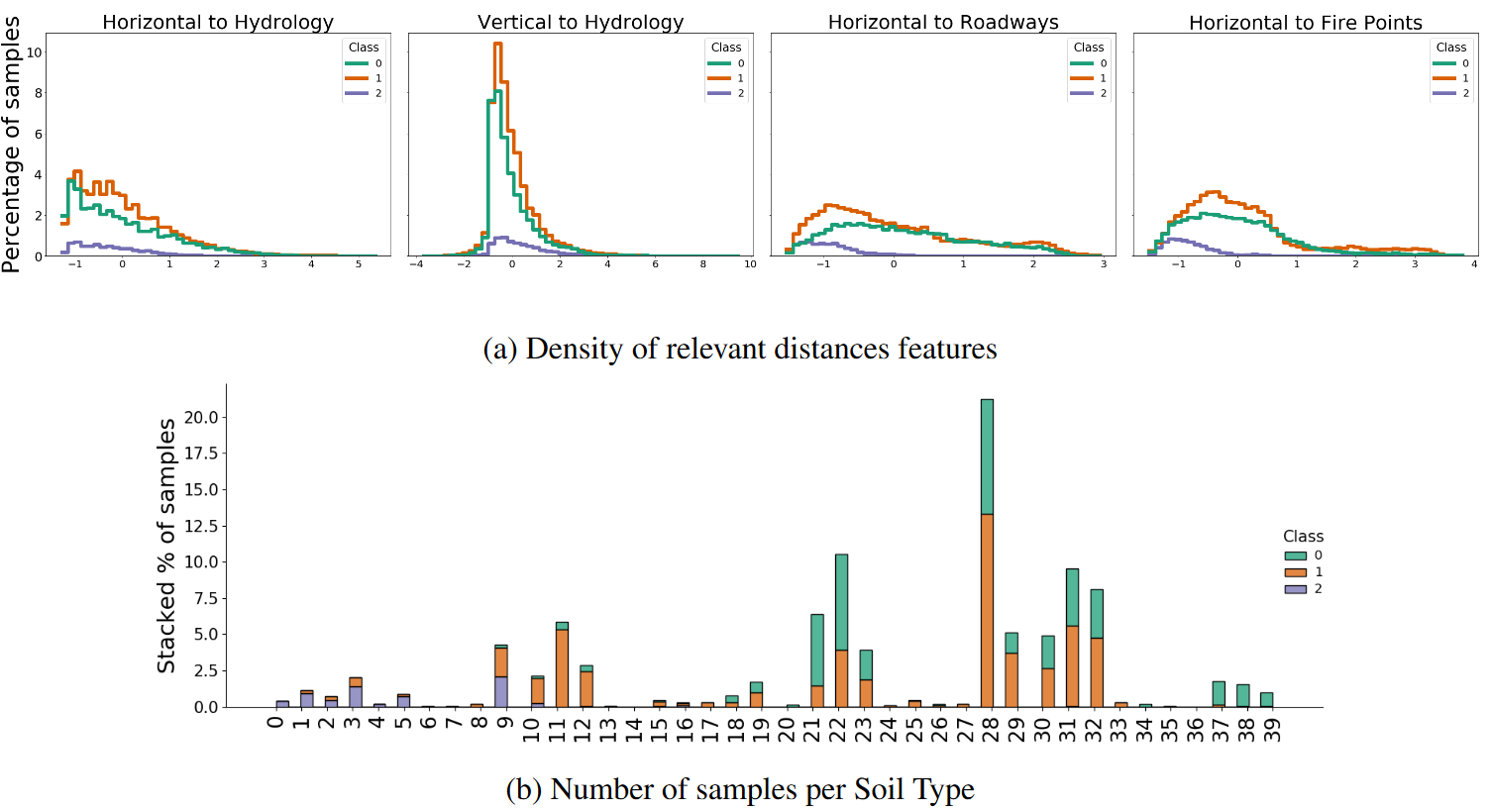}
\caption{CT's Exploratory Data Analysis.}
\label{figEDA_CT}
\end{figure*}

An Exploratory Data Analysis (EDA) was conducted on CT and NI as a means to validate which features are most relevant for each class. Notably, in Fig. \ref{figEDA_CT}a, we observe that CT's features corresponding to concept group \textit{Distances} do not seem to be particularly distinctive among classes according to their distributions. Perhaps their predictive power is better in conjunction with other groups. In contrast, Fig. \ref{figEDA_CT}b shows that \textit{Soil Type} does provide a clear differentiation between classes. All samples from Class $2$ have soil types in $\{0, ..., 9\}$, whereas samples from Class $0$ do not have soil types lower than $9$. Even though the EDA clearly shows \textit{Soil Type} concept group's relevance for the classification task, only LL and MLA methods capture this information.

In the NI dataset most features are continuous. Hence, only a small subset of them are presented and shown in Fig. \ref{fig8} in Appendix B. Through a similar EDA, we find that, akin to our distribution analysis, Class $1$ displays unexpected behaviors, with correspondences to the EDA appearing unclear. Nonetheless, both \textit{Host} and \textit{Traffic} emerge as strong indicators across all classes, which aligns with the majority of the distributions presented in Fig. \ref{figBestCGperClass-NI}.

In general, we conclude that LL and MLA are better aligned with the findings of the EDA. For Classes $0$ and $1$, both methods show similar correspondences -and disagreements- with it. However, for the least represented class in each dataset (Class $2$), MLA's aligment to the EDA appears better than LL's, as it focuses on the groups highlighted by the EDA for a larger number of samples.

\subsubsection*{Explanation Visualizations}
To get a visual representation of the explanations for a given sample, each of the compared methods' explainability values for each concept group are plotted in the heatmaps below. In the 2D heatmaps, for $j, k \in {1, ..., m}$, the values for MLA and AVG correspond to the probability of the maximum probability path $p$ between $v^{0}_{j}$ and $v^{M}_{k}$, whereas the values for LL correspond to $a^{M}_{j,k}$. Lighter color tones correspond to lower explainability values. For comparability across methods, values have been scaled to $[0, 1]$ and only correctly classified samples were considered. 

\begin{figure}[ht]
        \centering
        \subfloat[Sample 36, Class 1]{\includegraphics[width=0.45\textwidth]{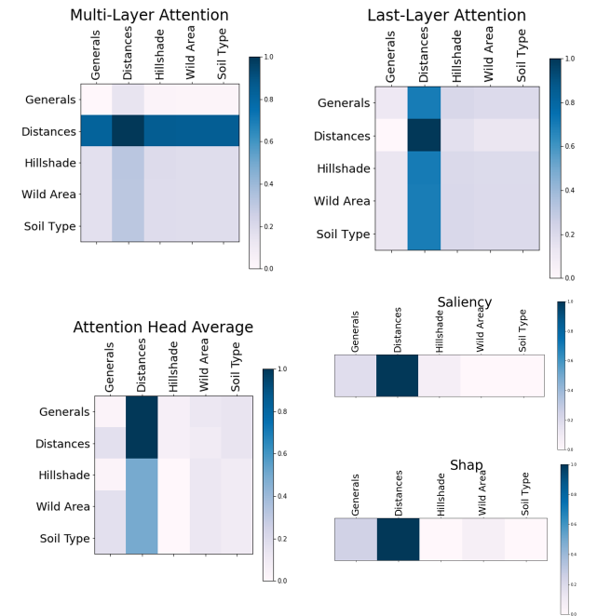}
        } \hspace{0.8cm}
        \subfloat[Sample 94, Class 2]{\includegraphics[width=0.43\textwidth]{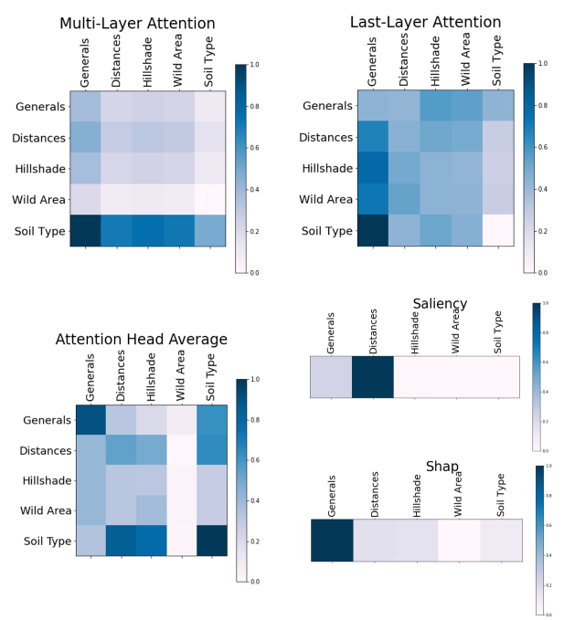}
        }
    \caption{CT Concept groups explainability coefficients.}  \label{figVis_CT}
\end{figure}

\begin{figure}[ht]
        \centering
        \subfloat[Sample 21259, Class 0]{\includegraphics[width=0.45\textwidth]{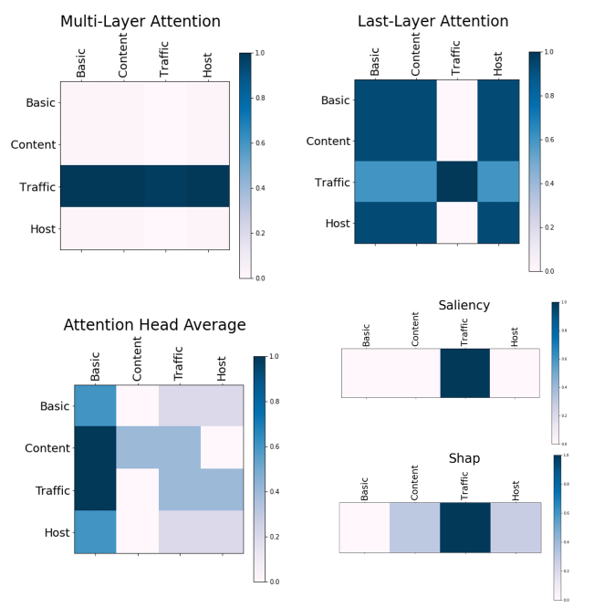}
        }\hspace{0.8cm}
        \subfloat[Sample 13927, Class 1]{\includegraphics[width=0.45\textwidth]{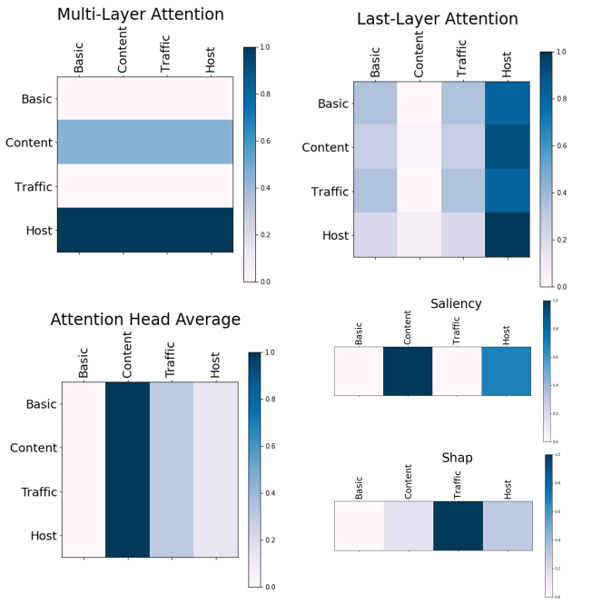}
        }
    \caption{NI Concept groups explainability coefficients..}  \label{figVis_NI}
\end{figure}

Fig. \ref{figVis_CT} and Fig. \ref{figVis_NI} show explanations corresponding to a couple of correctly classified samples from datasets CT and NI. Fig. \ref{figVis_CT}a shows a sample of CT where all methods identified \textit{Distances} as the \textit{best concept group}. This implies that the information obtained from the distances to hydrology, roadways, and fire points was identified by all methods as the most relevant for the model to conclude that the correct class was $1$. In contrast, in Fig. \ref{figVis_CT}b we observe a sample of Class $2$ for which methods LL, AVG and MLA identified \textit{Soil Type} as the \textit{best concept group}, whereas SA and SH assigned larger explainability values to \textit{Distances} and \textit{Generals}, respectively. 
Similarly, two NI samples are presented in Fig. \ref{figVis_NI}. In Fig. \ref{figVis_NI}a, we observe that features related to \textit{Traffic} were given larger values by all methods. In contrast, in Fig. \ref{figVis_NI}b there is a lack of unanimous agreement across methods, but with alignment shown by the attention-based methods.

\subsubsection*{Pairwise Method Comparison}
\begin{figure}[H]%
    \centering
    \subfloat[\centering CT]{{\includegraphics[width=0.33\textwidth]{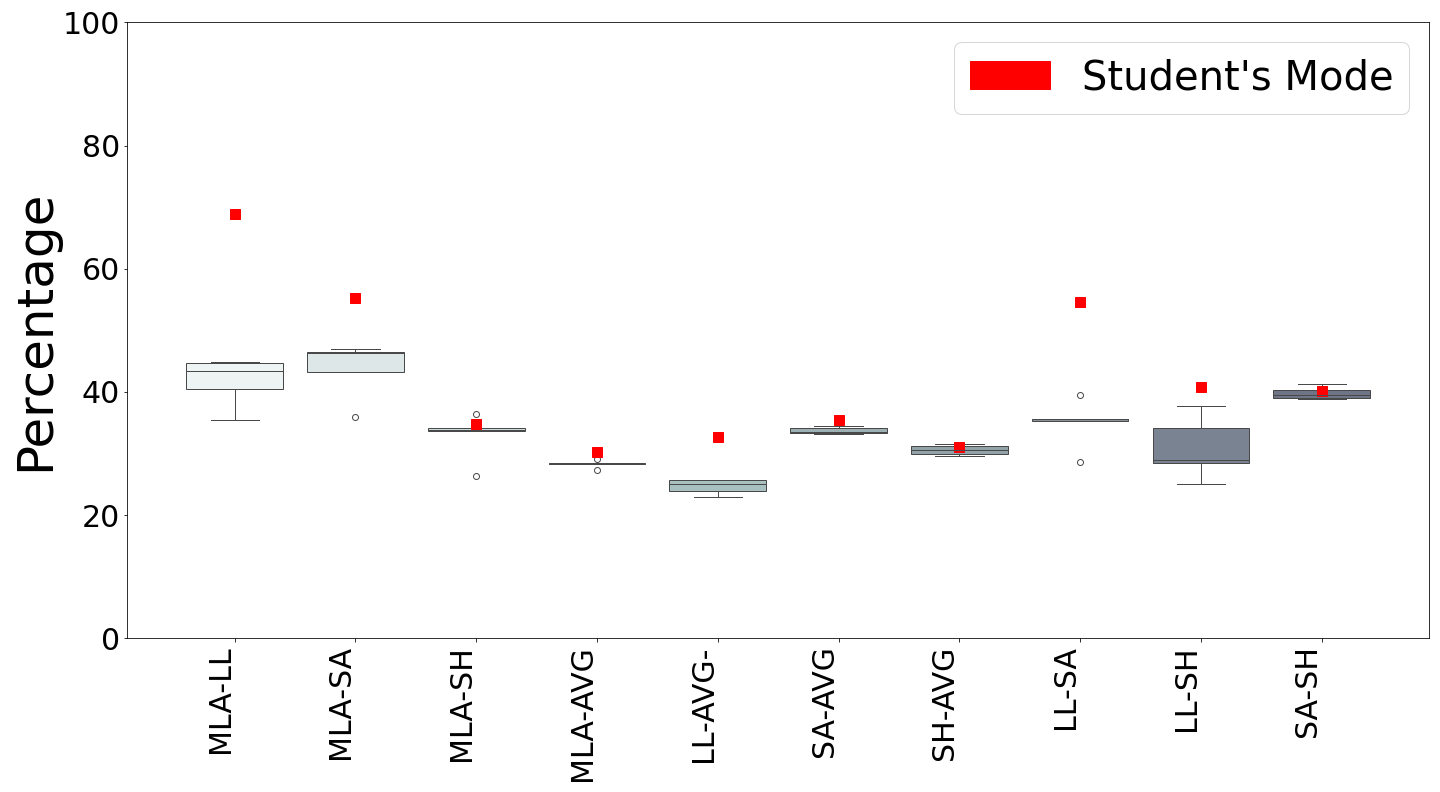} }}%
    \subfloat[\centering NI]{{\includegraphics[width=0.33\textwidth]
    {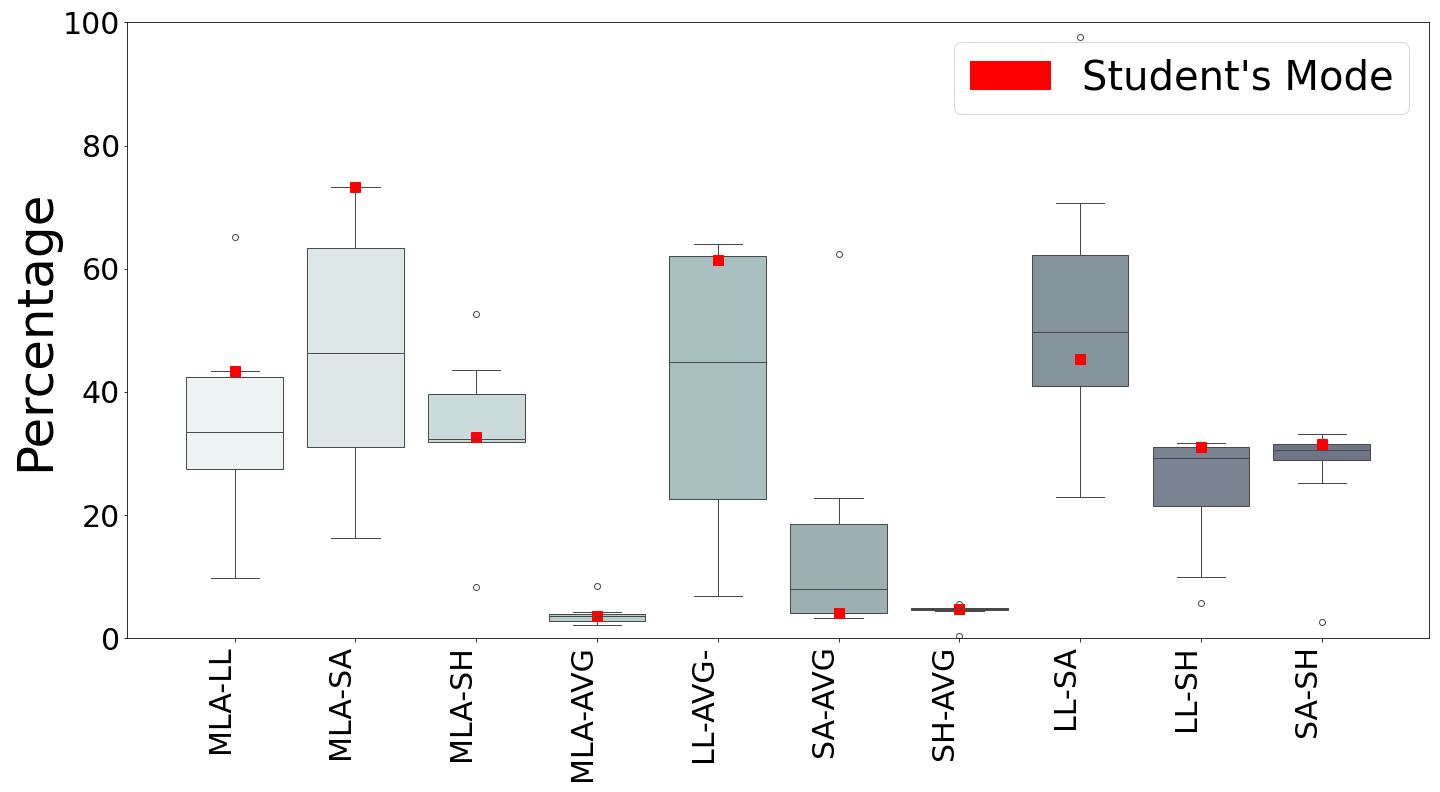} }}%
    \subfloat[\centering RW]{{\includegraphics[width=0.33\textwidth]{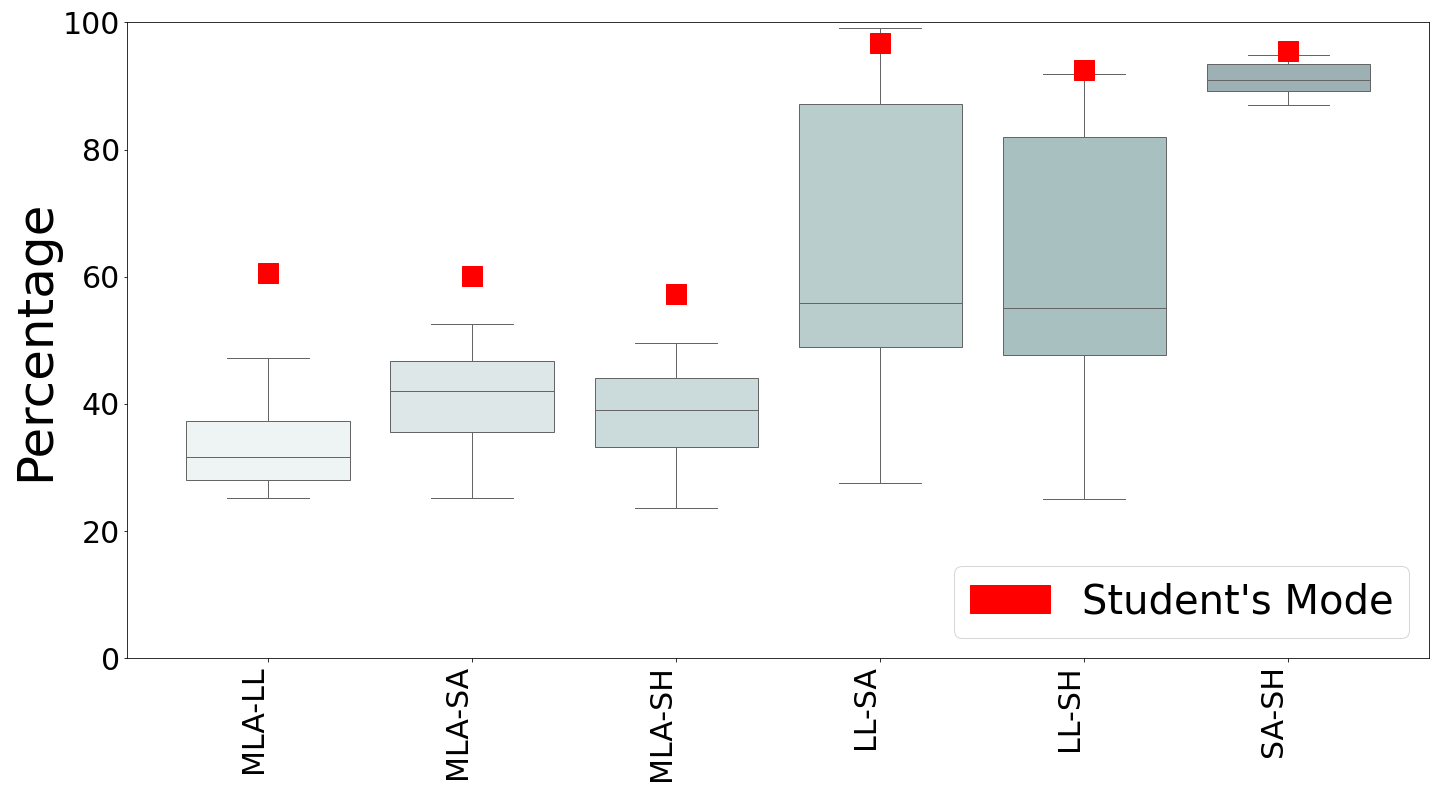} }}%
    \caption{Best context group pairwise comparison}%
    \label{figPairwise}%
\end{figure}

\begin{figure}[H]%
    \centering
    \subfloat[\centering CT]{{\includegraphics[width=0.34\textwidth]{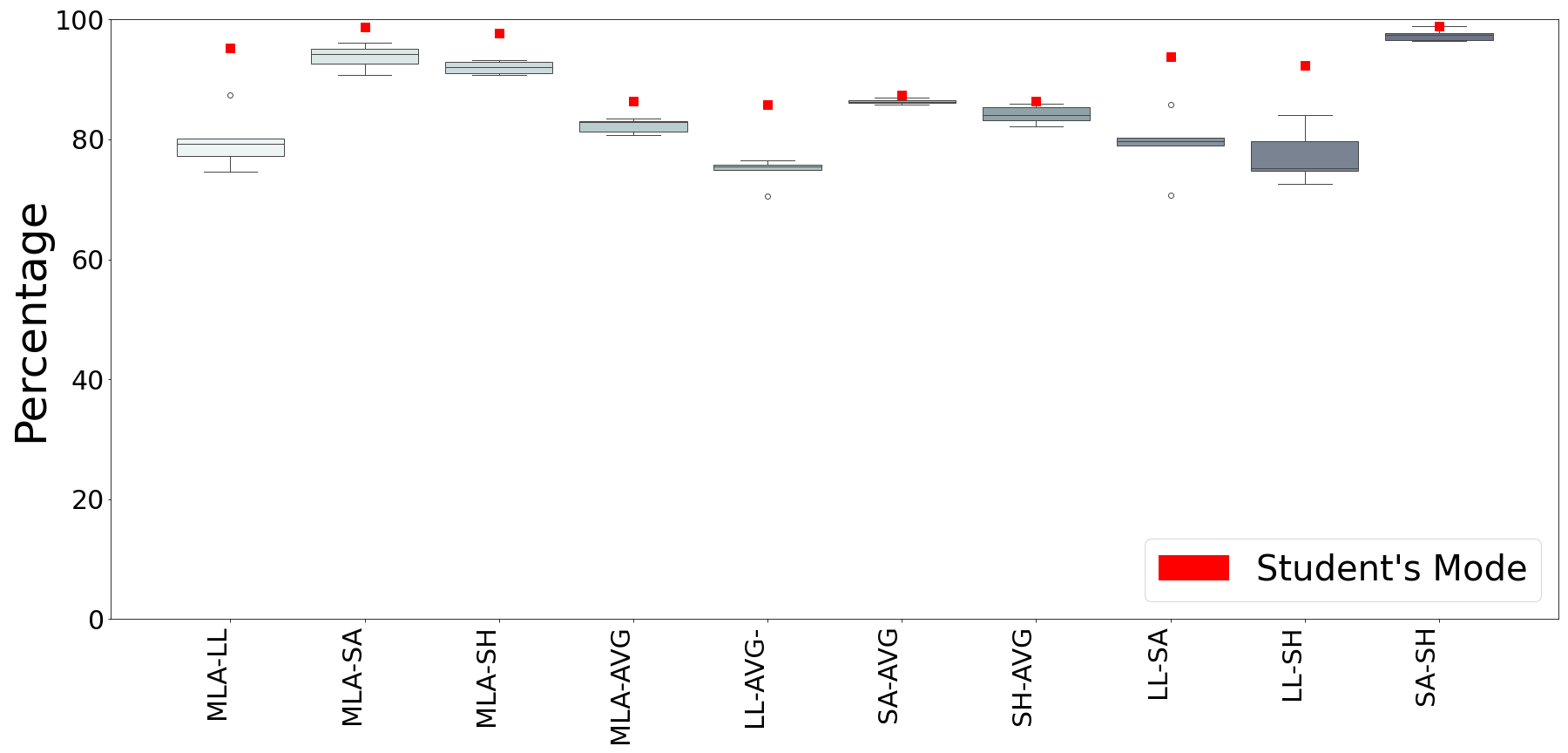} }}%
    \subfloat[\centering NI]{{\includegraphics[width=0.34\textwidth]{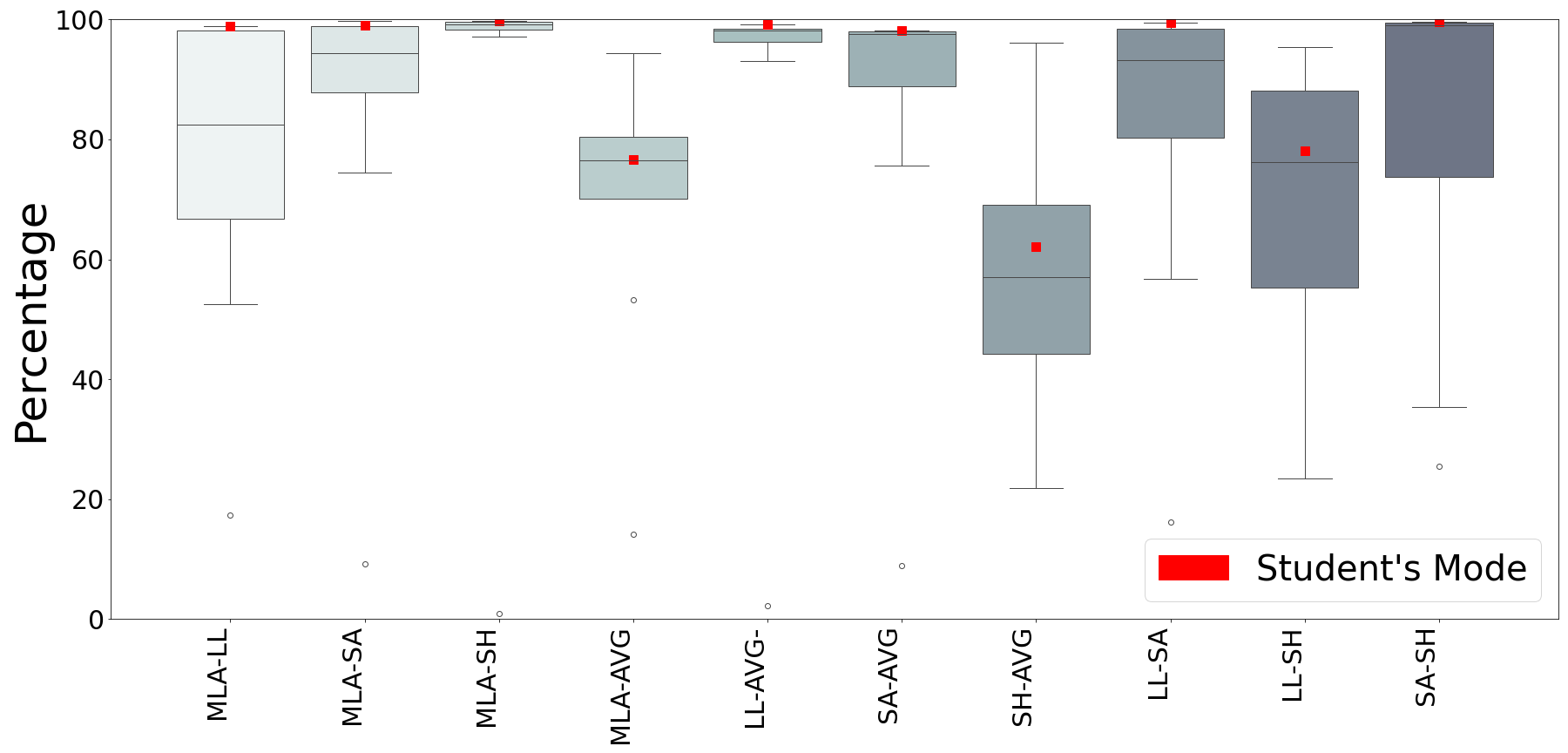} }}%
    \subfloat[\centering RW]{{\includegraphics[width=0.3\textwidth]{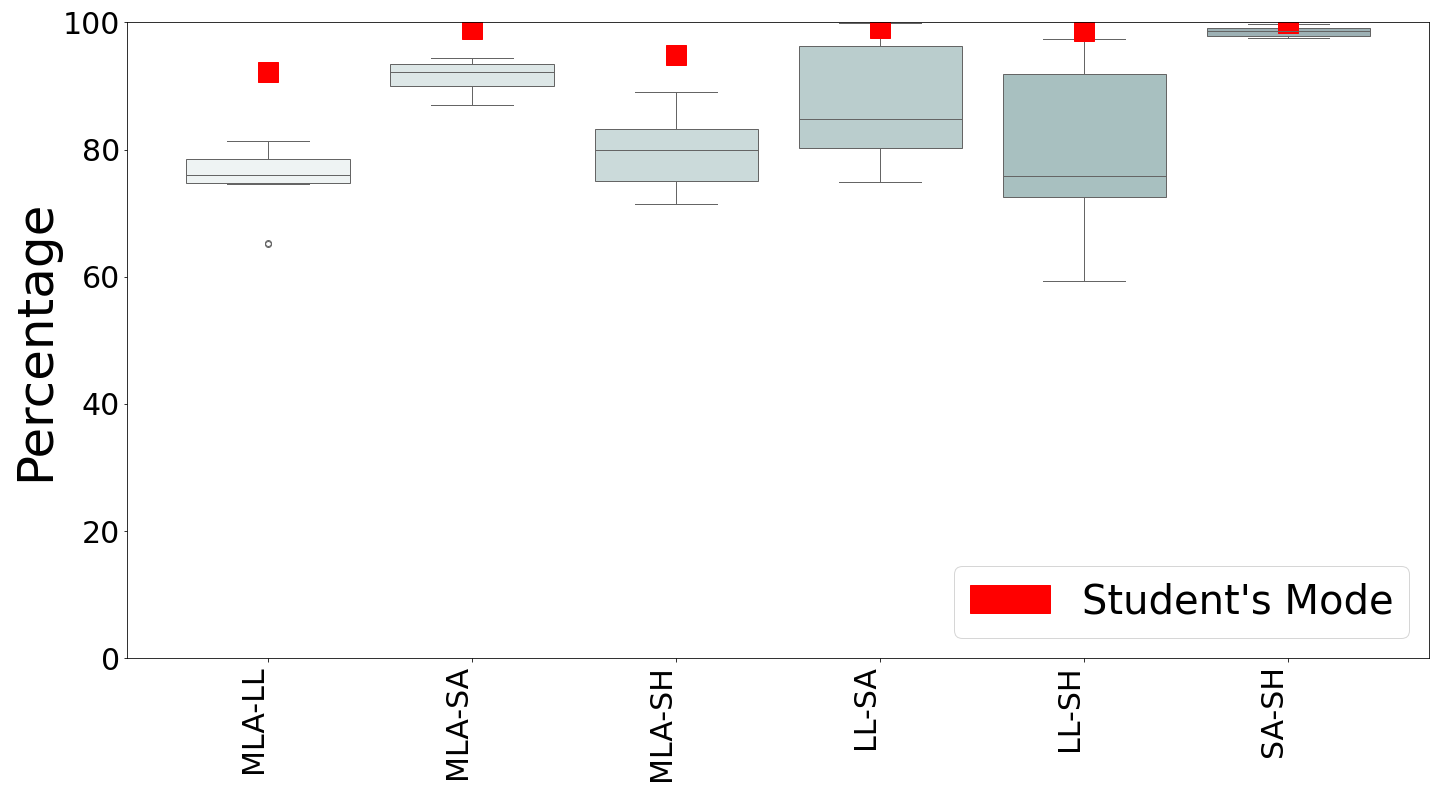} }}%
    \caption{Two best context groups pairwise comparison}%
    \label{figPairwise2B}%
\end{figure}

We now contrast the results provided per method by conducting pairwise comparisons among them. To do so, we quantify the number of samples for which the selected \textit{best concept group} is the same for two methods, i.e., for what percentage of the samples do two methods choose the same \textit{best concept group}. The distributions of such values across the various runs are presented in Fig. \ref{figPairwise}. On average, the pairwise comparisons with MLA are higher for CT and NI. The pairwise agreements of AVG with other methods exhibit notably lower values than the other methods. As seen in Fig. \ref{figPairwise}c, MLA seems to provide very different explanations to the ones generated by other methods for RW. The red square in each boxplot corresponds to the percentage of samples that chose the same \textit{best concept group} when defined as the mode across all repetitions. We expected MLA and LL to have high agreement, however, this is not always the case. As observed in the EDA, there are instances where a single group does not consistently provide all the relevant information for a model to predict a class for a given sample. To account for this, we consider the top-ranked concept groups. The number of concepts groups to include may vary depending on the application or could be determined by setting a predefined probability threshold. For simplicity, we select the top \textit{two best concept groups} per method and quantify the percentage of samples where at least one of them is common to each pair of methods. The resulting distributions, means, and modes per pairwise comparison are reported in Fig. \ref{figPairwise2B}.

When considering the \textit{two best concept groups}, we observe very high pairwise agreement across methods. For MLA, all mean and mode pairwise overlaps are above $75\%$ and $92\%$, respectively. MLA, SA, and SH consistently show large pairwise agreements across datasets, while LL and AVG yield the smallest number of coinciding explanations.
Overall, MLA exhibits similarities to LL, as anticipated given their shared utilization of attention-based approaches. However, these similarities do not extend to AVG, despite its close resemblance to MLA in terms of methodology. Across all methods, LL displays the highest variability, a tendency that appears to be alleviated by MLA's comprehensive consideration of attention graphs. Despite AVG's inclusion of attention graphs, its performance appears markedly distinct to LL and MLA, likely attributable to a potential loss of information resulting from the multi-head averaging operation. Additionally, MLA shows better pairwise results with SA and SH than LL and AVG.

On the other hand, our experiments show that gradient- and perturbation-based methods (SA and SH) are more similar to each other than to the attention-based methods. They produce similar explanations and focus mostly on a reduced number of groups (which are not necessarily different across classes) to generate explanations for predictions. 

\subsubsection{Stability Analysis}
\begin{figure}[h]%
    \centering
    \subfloat[\centering CT]{\includegraphics[width=0.25\linewidth]{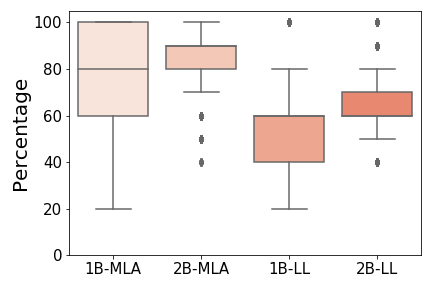}}%
    \subfloat[\centering NI]{{\includegraphics[width=0.25\linewidth]{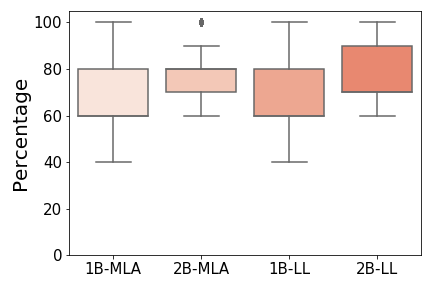} }}%
    \subfloat[\centering RW]{{\includegraphics[width=0.25\linewidth]{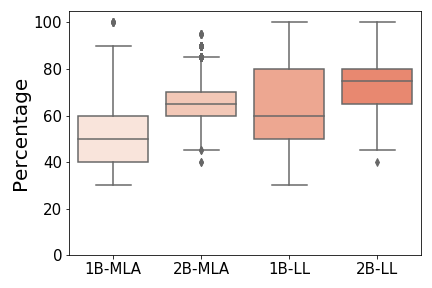} }}%
    \caption{Percentage of runs that agree on the best (1B) and two best (2B) \textit{context groups} per method}%
    \label{figStability}%
\end{figure}

The stability of the explanations is analyzed by quantifying the percentage of distinct runs\footnote{Note that one teacher network was trained for each dataset, and its corresponding student networks underwent multiple runs to accommodate randomization. Since AVG is derived from the teacher network rather than the student networks, it is excluded from this analysis.} that agree on the same explanation for each sample. Given the previously discussed observation that SA and SH tend to steadily choose the same groups even across different samples, we focus on methods MLA and LL for this analysis. Fig. \ref{figStability} shows the boxplots for the best (1B) and two best (2B) \textit{concept groups} per dataset.

For these datasets, the 1B \textit{concept groups} comparison of MLA and LL appears inconclusive. For CT, we observe a better performance of MLA but larger variability than LL. On the other hand, the exact opposite can be said for RW, whereas both distributions seem to be identical for NI. It is important to note that correlation between concept groups could have a major impact in the 1B results. In the extreme case in which the data has perfectly correlated groups, the methods are free to choose one group over the other at random. Identifying the \textit{two best concept groups} helps to mitigate this issue. Our experiments using the 2B \textit{concept groups} show that both models are quite stable with averages of over $60\%$ of agreement across runs. Again, the average model-to-model comparison seems to be dataset-dependant, however, MLA consistently shows lower variability than LL, making it more reliable and prone to provide robust and reliable explanations.

\section{Conclusion} \label{sec:conclusion}
In this paper, we present a novel explainability method for TD that leverages transformer models and incorporates knowledge from the graph structure of attention matrices. Combining these two, we propose a way of identifying the concept groups of input features that provide the model with the most relevant information to make a prediction. We compare our method with well-known gradient-, attention-, and perturbation-based explanations and highlight the similarities and dissimilarities observed in our experiments. 

\bibliographystyle{abbrv}  
\bibliography{main}  

\begin{thebibliography}{10}

\bibitem{abnar}
S.~Abnar and W.~Zuidema.
\newblock Quantifying attention flow in transformers.
\newblock In {\em Proceedings of the 58th Annual Meeting of the Association for Computational Linguistics}. Association for Computational Linguistics, 2020.

\bibitem{NI-ds-groups}
P.~Aggarwal and S.~K. Sharma.
\newblock {Analysis of KDD dataset attributes - Class wise for intrusion detection}.
\newblock {\em Procedia Computer Science. International Conference on Recent Trends in Computing}, 2015.

\bibitem{tabnet2019}
S.~O. Arik and T.~Pfister.
\newblock Tabnet: Attentive interpretable tabular learning.
\newblock In {\em Proceedings of the AAAI Conference on Artificial Intelligence}, 2021.

\bibitem{LWRP}
S.~Bach, A.~Binder, G.~Montavon, F.~Klauschen, K.-R. M{\"u}ller, and W.~Samek.
\newblock On pixel-wise explanations for non-linear classifier decisions by layer-wise relevance propagation.
\newblock {\em PLOS ONE}, 2015.

\bibitem{Blackard:1999}
J.~A. Blackard.
\newblock {UCI Machine Learning Repository}, 1999.
\newblock \url{https:// archive.ics.uci.edu/ml/machine-learning-databases/covtype/ covtype.info}.

\bibitem{survey}
V.~Borisov, T.~Leemann, K.~Seßler, J.~Haug, M.~Pawelczyk, and G.~Kasneci.
\newblock Deep neural networks and tabular data: A survey.
\newblock {\em arXiv preprint arXiv:2110.01889}, 2021.

\bibitem{detr}
N.~Carion, F.~Massa, G.~Synnaeve, N.~Usunier, A.~Kirillov, and S.~Zagoruyko.
\newblock End-to-end object detection with transformers.
\newblock In {\em European Conference on Computer Vision}, 2020.

\bibitem{caruana2015}
R.~Caruana, Y.~Lou, J.~Gehrke, P.~Koch, M.~Sturm, and N.~Elhadad.
\newblock Intelligible models for healthcare: Predicting pneumonia risk and hospital 30-day readmission.
\newblock In {\em Proceedings of the International Conference on Knowledge Discovery and Data Mining}, 2015.

\bibitem{chefer}
H.~Chefer, S.~Gur, and L.~Wolf.
\newblock Transformer interpretability beyond attention visualization.
\newblock In {\em {IEEE} Conference on Computer Vision and Pattern Recognition}, 2021.

\bibitem{xgboost}
T.~Chen and C.~Guestrin.
\newblock {XGBoost}: A scalable tree boosting system.
\newblock In {\em Proceedings of the 22nd ACM SIGKDD International Conference on Knowledge Discovery and Data Mining}, KDD '16, pages 785--794, New York, NY, USA, 2016. ACM.

\bibitem{confalonieri2021}
R.~Confalonieri, L.~Coba, B.~Wagner, and T.~Besold.
\newblock A historical perspective of explainable artificial intelligence.
\newblock {\em Wiley Interdisciplinary Reviews: Data Mining and Knowledge Discovery}, 2021.

\bibitem{das2020}
A.~Das and P.~Rad.
\newblock Opportunities and challenges in explainable artificial intelligence {(XAI):} {A} survey.
\newblock {\em arXiv preprint arXiv:2006.11371}, 2020.

\bibitem{dijkstra1959note}
E.~W. Dijkstra.
\newblock A note on two problems in connexion with graphs.
\newblock {\em Numerische mathematik}, 1959.

\bibitem{dong}
Y.~Dong, J.-B. Cordonnier, and A.~Loukas.
\newblock Attention is not all you need: pure attention loses rank doubly exponentially with depth.
\newblock In {\em Proceedings of the 38th International Conference on Machine Learning}, 2021.

\bibitem{lopardo}
L.~et. al.
\newblock Attention meets post-hoc interpretability: A mathematical perspective, 2024.

\bibitem{mylonas}
M.~et. al.
\newblock An attention matrix for every decision: faithfulness-based arbitration among multiple attention-based interpretations of transformers in text classification.
\newblock {\em Data Mining and Knowledge Discovery}, 2024.

\bibitem{niu}
N.~et. al.
\newblock Attexplainer: Explain transformer via attention by reinforcement learning.
\newblock In {\em Proceedings of the Thirty-First International Joint Conference on Artificial Intelligence}, 2022.

\bibitem{Goodfellow-et-al-2016}
I.~Goodfellow, Y.~Bengio, and A.~Courville.
\newblock {\em Deep Learning}.
\newblock MIT Press, 2016.
\newblock \url{http://www.deeplearningbook.org}.

\bibitem{fttransf2021}
Y.~Gorishniy, I.~Rubachev, V.~Khrulkov, and A.~Babenko.
\newblock Revisiting deep learning models for tabular data.
\newblock In {\em Proceedings of the International Conference on Advances in Neural Information Processing Systems}, 2021.

\bibitem{gunning2019}
D.~Gunning, M.~Stefik, J.~Choi, T.~Miller, S.~Stumpf, and G.-Z. Yang.
\newblock {XAI—Explainable artificial intelligence}.
\newblock {\em Science Robotics}, 2019.

\bibitem{NI-ds}
S.~Hettich and S.~D. Bay.
\newblock {UCI KDD Archive}, 1999.
\newblock \url{http://kdd.ics.uci.edu/databases/kddcup99/kddcup99.html}.

\bibitem{hinton2015distilling}
G.~Hinton, O.~Vinyals, and J.~Dean.
\newblock Distilling the knowledge in a neural network.
\newblock {\em International Conference on Advances in Neural Information Processing Systems - Deep Learning and Representation Learning Workshop}, 2014.

\bibitem{tabtransf2020}
X.~Huang, A.~Khetan, M.~Cvitkovic, and Z.~S. Karnin.
\newblock Tabular data modeling using contextual embeddings.
\newblock {\em 9th International Conference on Learning Representations - Workshop on Weakly Supervised Learning}, 2021.

\bibitem{islam}
S.~R. Islam, W.~Eberle, S.~K. Ghafoor, and M.~Ahmed.
\newblock Explainable artificial intelligence approaches: {A} survey.
\newblock {\em arXiv preprint arXiv:2101.09429}, 2021.

\bibitem{jain}
S.~Jain and B.~C. Wallace.
\newblock {A}ttention is not {E}xplanation.
\newblock In {\em Proceedings of the 2019 Conference of the North {A}merican Chapter of the Association for Computational Linguistics: Human Language Technologies}, 2019.

\bibitem{lightgbm}
G.~Ke, Q.~Meng, T.~Finley, T.~Wang, W.~Chen, W.~Ma, Q.~Ye, and T.-Y. Liu.
\newblock Lightgbm: A highly efficient gradient boosting decision tree.
\newblock {\em Advances in neural information processing systems}, 30:3146--3154, 2017.

\bibitem{adam}
D.~P. Kingma and J.~Ba.
\newblock Adam: A method for stochastic optimization.
\newblock {\em 3rd International Conference for Learning Representations}, 2015.

\bibitem{transformersurvey}
T.~Lin, Y.~Wang, X.~Liu, and X.~Qiu.
\newblock A survey of transformers.
\newblock {\em AI Open}, 2022.

\bibitem{vilbert}
J.~Lu, D.~Batra, D.~Parikh, and S.~Lee.
\newblock Vi{LBERT}: Pretraining task-agnostic visiolinguistic representations for vision-and-language tasks.
\newblock In {\em Proceedings of the International Conference on Advances in Neural Information Processing Systems}, 2019.

\bibitem{SHAP}
S.~M. Lundberg and S.~I. Lee.
\newblock A unified approach to interpreting model predictions.
\newblock In {\em Proceedings of the International Conference on Advances in Neural Information Processing Systems}, 2017.

\bibitem{LIME}
M.~Ribeiro, S.~Singh, and C.~Guestrin.
\newblock {``}{W}hy should {I} trust you?{''}: Explaining the predictions of any classifier.
\newblock In {\em Proceedings of the Conference of the North {A}merican Chapter of the Association for Computational Linguistics: Demonstrations}, 2016.

\bibitem{Roy2019}
S.~Roy, V.~Balas, P.~Samui, and S.~D.
\newblock {\em Handbook of Deep Learning applications}.
\newblock Springer, 2019.

\bibitem{samek2019explainable}
W.~Samek, G.~Montavon, A.~Vedaldi, L.~Hansen, and K.~M{\"u}ller.
\newblock {\em Explainable AI: Interpreting, explaining and visualizing Deep Learning}.
\newblock Lecture Notes in Computer Science. Springer, 2019.

\bibitem{gradcam}
R.~R. Selvaraju, M.~Cogswell, A.~Das, R.~Vedantam, D.~Parikh, and D.~Batra.
\newblock Grad-cam: Visual explanations from deep networks via gradient-based localization.
\newblock In {\em IEEE International Conference on Computer Vision}, 2017.

\bibitem{serrano}
S.~Serrano and N.~A. Smith.
\newblock Is attention interpretable?
\newblock In {\em Proceedings of the 57th Annual Meeting of the Association for Computational Linguistics}, 2019.

\bibitem{deeplift}
A.~Shrikumar, P.~Greenside, and A.~Kundaje.
\newblock Learning important features through propagating activation differences.
\newblock In {\em Proceedings of the International Conference on Machine Learning}, 2017.

\bibitem{Simonyan}
K.~Simonyan, A.~Vedaldi, and A.~Zisserman.
\newblock Deep inside convolutional networks: Visualising image classification models and saliency maps.
\newblock {\em Workshop at International Conference on Learning Representations}, 2014.

\bibitem{smoothgrad}
D.~Smilkov, N.~Thorat, B.~Kim, F.~B. Vi{\'{e}}gas, and M.~Wattenberg.
\newblock Smooth{G}rad: {R}emoving noise by adding noise.
\newblock {\em International Conference on Machine Learning - Workshop on Visualization for Deep Learning}, 2017.

\bibitem{saint2021}
G.~Somepalli, M.~Goldblum, A.~Schwarzschild, C.~B. Bruss, and T.~Goldstein.
\newblock {SAINT:} {I}mproved neural networks for tabular data via row attention and contrastive pre-training.
\newblock {\em arXiv preprint arXiv:2106.01342}, 2021.

\bibitem{autoint2019}
W.~Song, C.~Shi, Z.~Xiao, Z.~Duan, Y.~Xu, M.~Zhang, and J.~Tang.
\newblock Autoint: Automatic feature interaction learning via self-attentive neural networks.
\newblock In {\em Proceedings of the ACM International Conference on Information and Knowledge Management}, 2019.

\bibitem{lxmert}
H.~Tan and M.~Bansal.
\newblock {LXMERT:} {L}earning cross-modality encoder representations from transformers.
\newblock In {\em Proceedings of the Conference on Empirical Methods in Natural Language Processing and the 9th International Joint Conference on Natural Language Processing}, 2019.

\bibitem{thorne}
J.~Thorne, A.~Vlachos, C.~Christodoulopoulos, and A.~Mittal.
\newblock Generating token-level explanations for natural language inference.
\newblock In {\em Proceedings of the 2019 Conference of the North {A}merican Chapter of the Association for Computational Linguistics: Human Language Technologies}, 2019.

\bibitem{tjoa2019}
E.~Tjoa and C.~Guan.
\newblock A survey on explainable artificial intelligence {(XAI):} {T}owards medical {XAI}.
\newblock {\em IEEE Transactions on Neural Networks and Learning Systems}, 2021.

\bibitem{attention2017}
A.~Vaswani, N.~Shazeer, N.~Parmar, J.~Uszkoreit, L.~Jones, A.~N. Gomez, L.~u. Kaiser, and I.~Polosukhin.
\newblock Attention is all you need.
\newblock In {\em Proceedings of the International Conference on Advances in Neural Information Processing Systems}, 2017.

\bibitem{velickovic2018graph}
P.~Veli{\v{c}}kovi{\'{c}}, G.~Cucurull, A.~Casanova, A.~Romero, P.~Li{\`{o}}, and Y.~Bengio.
\newblock Graph attention networks.
\newblock {\em International Conference on Learning Representations}, 2018.

\bibitem{vilone}
G.~Vilone and L.~Longo.
\newblock Explainable artificial intelligence: {A} systematic review.
\newblock {\em arXiv preprint arXiv:2006.00093}, 2020.

\bibitem{notions}
G.~Vilone and L.~Longo.
\newblock Notions of explainability and evaluation approaches for explainable artificial intelligence.
\newblock {\em Information Fusion}, 2021.

\bibitem{voita}
E.~Voita, D.~Talbot, F.~Moiseev, R.~Sennrich, and I.~Titov.
\newblock Analyzing multi-head self-attention: Specialized heads do the heavy lifting, the rest can be pruned.
\newblock In {\em Proceedings of the Annual Meeting of the Association for Computational Linguistics}, 2019.

\bibitem{wiegreffe}
S.~Wiegreffe and Y.~Pinter.
\newblock Attention is not not explanation.
\newblock In {\em Proceedings of the 2019 Conference on Empirical Methods in Natural Language Processing and the 9th International Joint Conference on Natural Language Processing}, 2019.

\bibitem{visualsurvey}
Q.~Zhang and S.~Zhu.
\newblock Visual interpretability for deep learning: {A} survey.
\newblock {\em Frontiers of Information Technology \& Electronic Engineering}, 2018.

\end{thebibliography}
\include{subgraphics}

\onecolumn

\section*{Appendix A) Best Context Group Distributions by Sample Classification Output}

\begin{figure} [H]
    \centering
\subfloat{{\includegraphics[height=0.205\textwidth,width=0.33\linewidth]{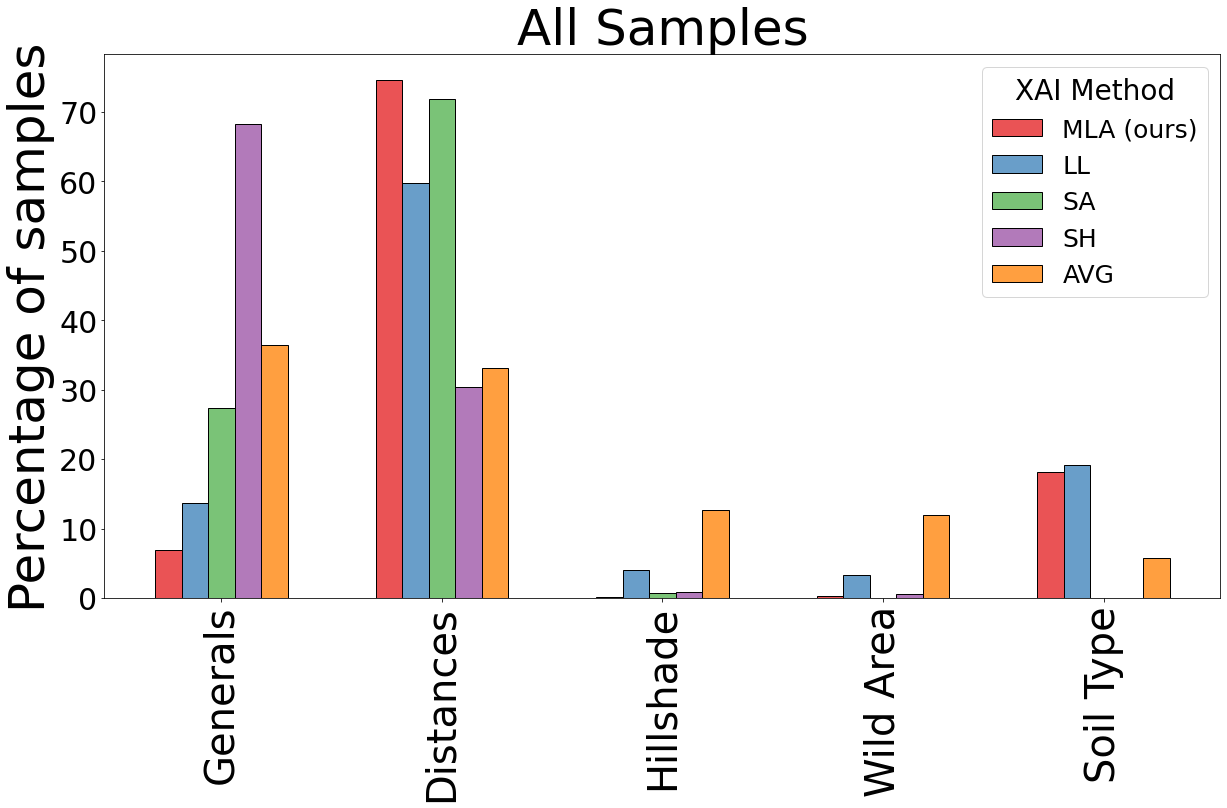} }}%
\subfloat{{\includegraphics[height=0.2\textwidth,width=0.315\linewidth]{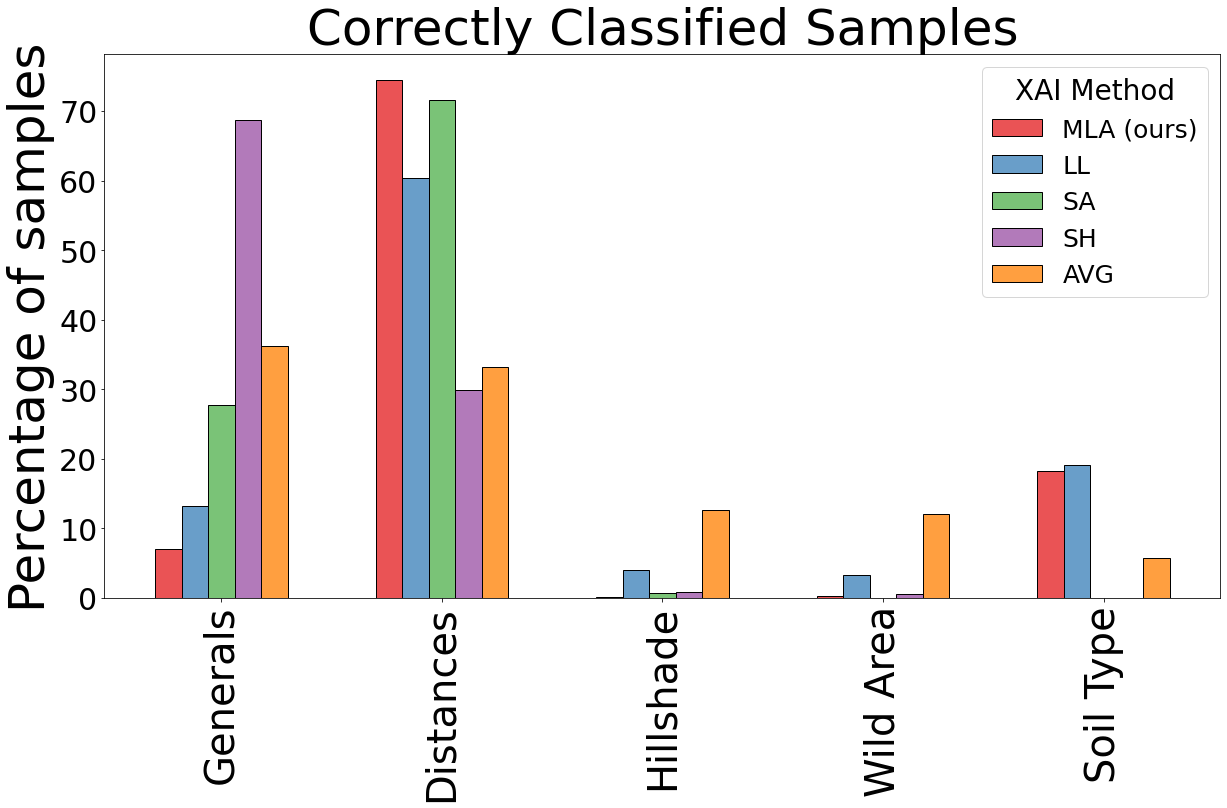} }}%
\subfloat{{\includegraphics[height=0.2\textwidth,width=0.315\linewidth]{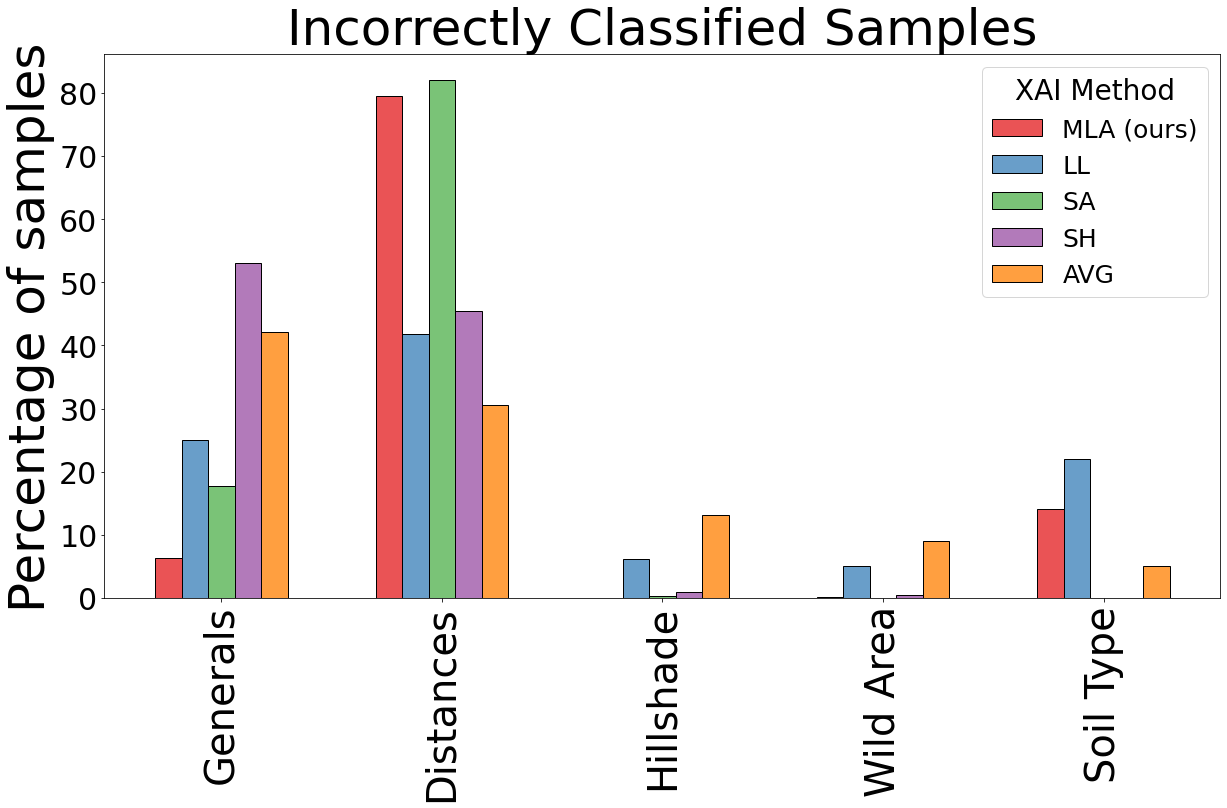} }}%
    \caption{CT}%
    \label{figCTsampletype}
\end{figure}

\begin{figure} [H]
    \centering
\subfloat{{\includegraphics[height=0.205\textwidth,width=0.33\linewidth]{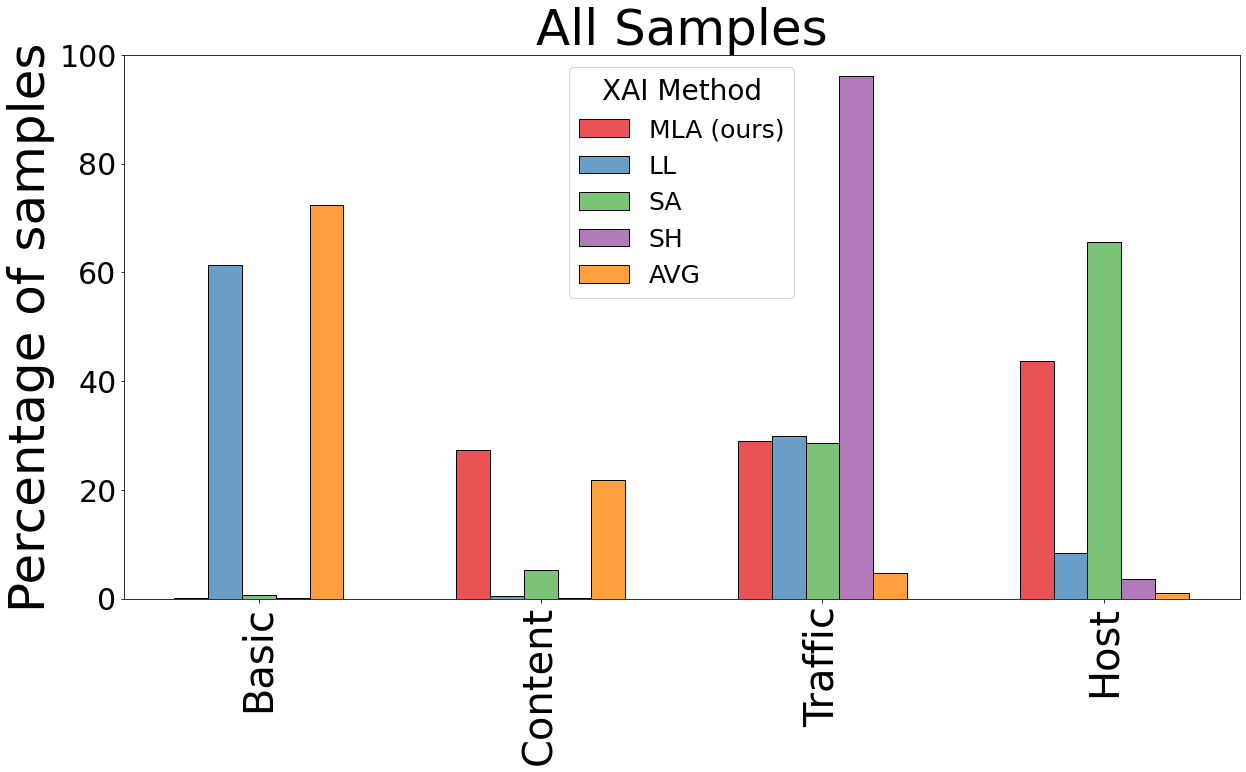} }}%
\subfloat{{\includegraphics[height=0.2\textwidth,width=0.315\linewidth]{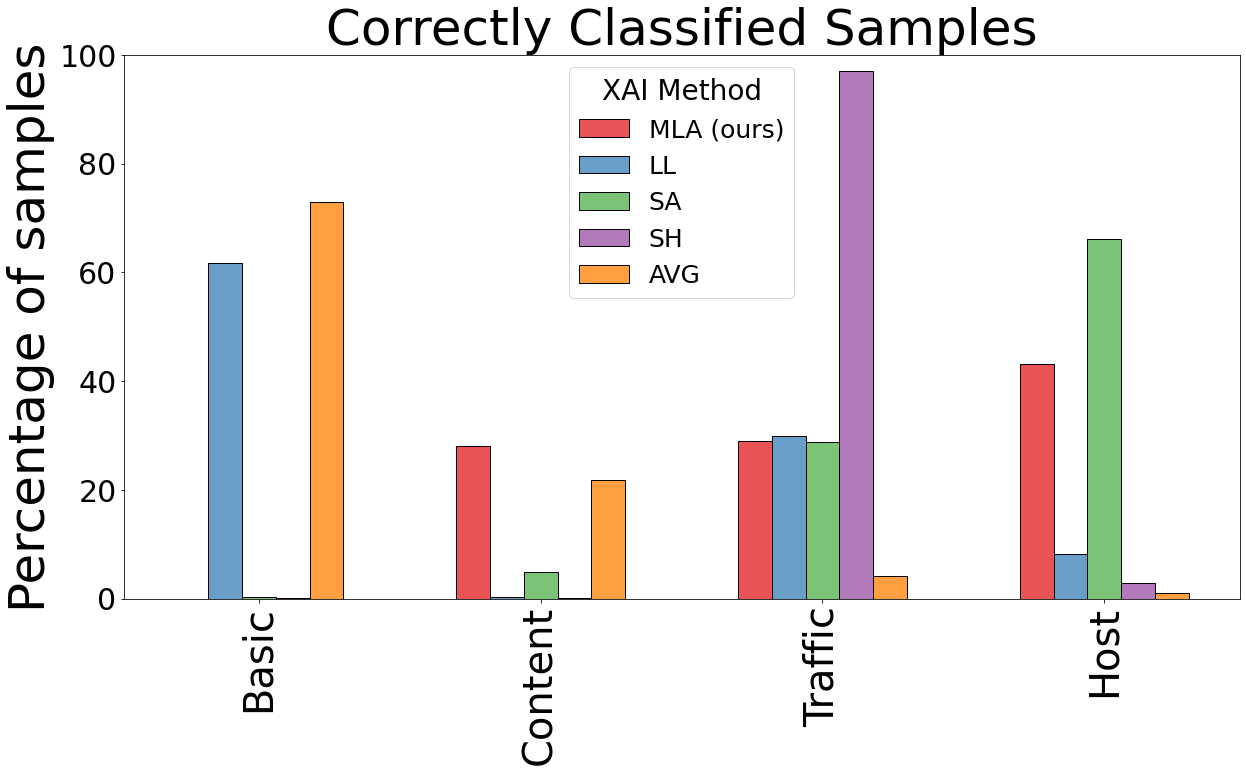} }}%
\subfloat{{\includegraphics[height=0.2\textwidth,width=0.315\linewidth]{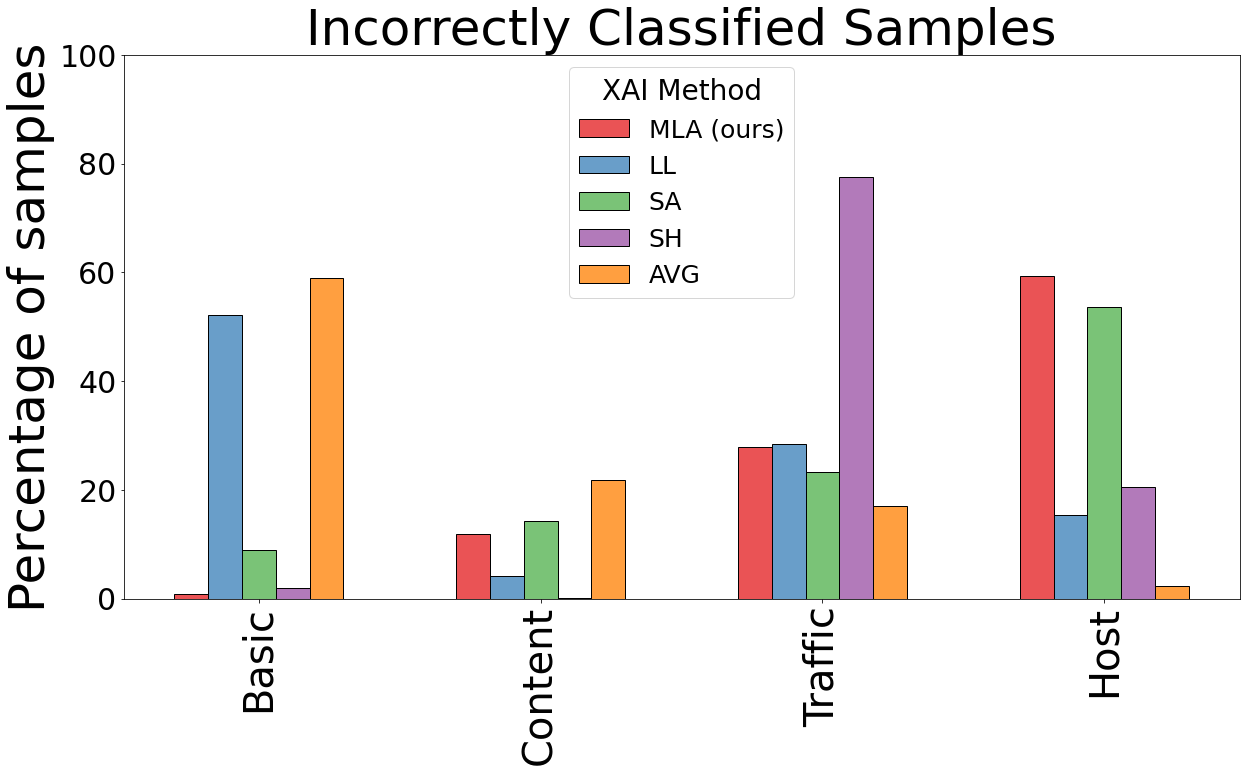} }}%
    \caption{NI}%
    \label{figNIsampletype}
\end{figure}

\vspace{10cm}

\pagebreak

\section*{Appendix B) Network Intrusion - Exploratory Data Analysis}

\begin{figure*}[h]
    \centering
    \includegraphics[width=0.8\linewidth]{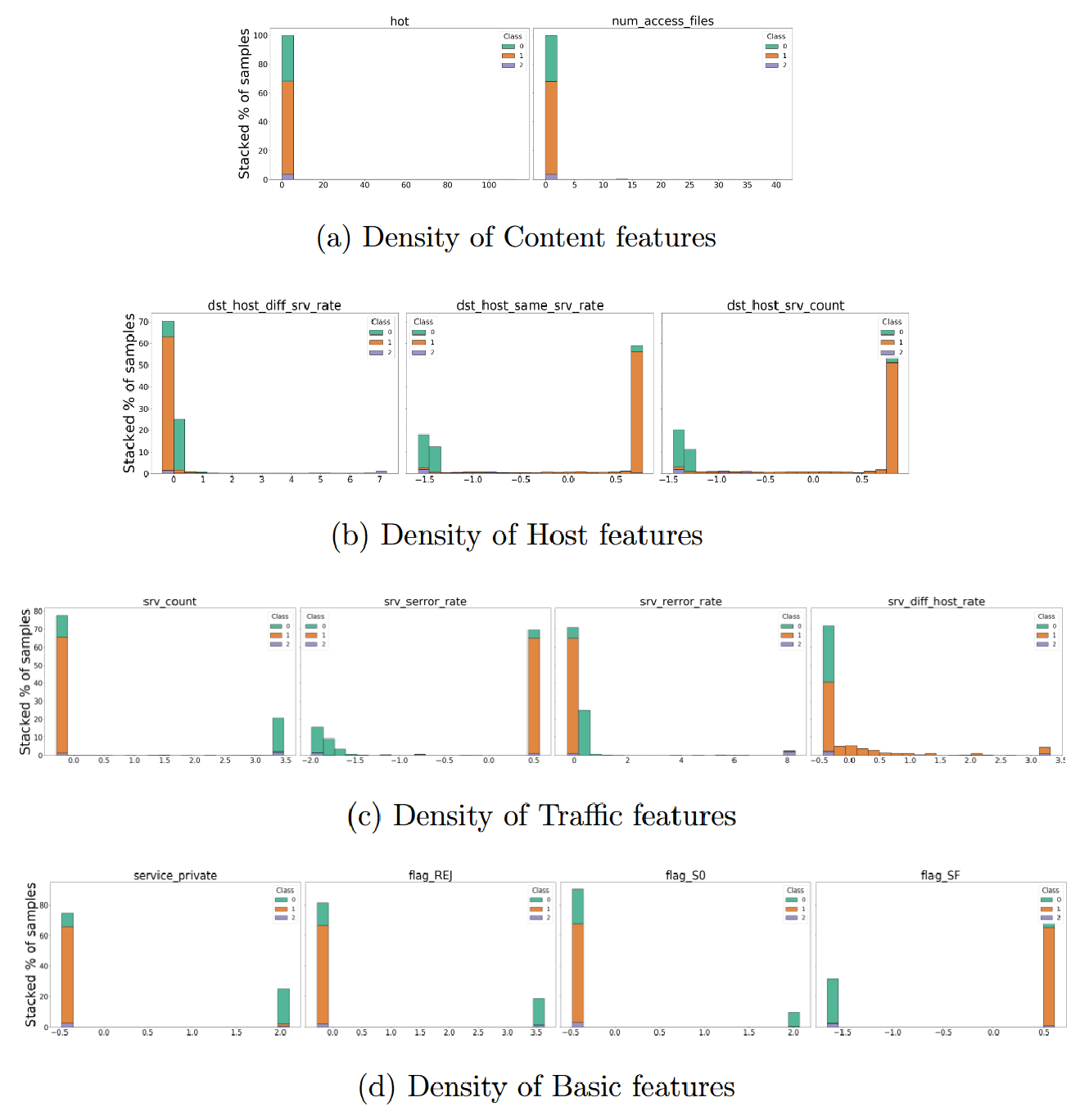}
\caption{NI Exploratory Data Analysis}
\label{fig8}
\end{figure*}

\end{document}